%% file: main.tex
\documentclass[11pt]{article}
\usepackage{latexsym}
\usepackage{amsmath}
\usepackage{amssymb}
\usepackage{amsthm}
\usepackage{epsfig}
\usepackage{xcolor}
\usepackage{graphicx}
\usepackage{bm}
\usepackage{enumitem}
\usepackage{mathtools}
\usepackage{graphicx}
\usepackage{tikz}
\usepackage{ mathrsfs }
\usepackage[toc,page]{appendix}
\usepackage[most]{tcolorbox}
\usepackage{graphicx}
\usepackage{bbm}
\usepackage{ytableau}
\usepackage{tkz-berge}
\usepackage{booktabs}
\usepackage{graphicx}
\usepackage{epstopdf}
\usepackage{titletoc}
\usepackage{array}
\usepackage{float}
\usepackage{booktabs}
\usepackage{wrapfig}
\usepackage[top=1in, bottom=1in, left=1in, right=1in]{geometry}

\usepackage{hyperref}
\hypersetup{
    colorlinks,
    linkcolor={blue!80!black},
    citecolor={green!50!black},
}
\colorlet{linkequation}{blue}
\usepackage{cleveref}
\crefname{appendix}{Appendix}{Appendices}
\Crefname{appendix}{Appendix}{Appendices}
\usepackage{authblk}
\usepackage{float}         \usepackage{tabularx}
\usepackage{algorithm}  
\usepackage{graphicx}
\usepackage{subcaption}
\usepackage[round,authoryear]{natbib}
\Crefname{algorithm}{Algorithm}{Algorithms}
\crefname{algorithm}{algorithm}{algorithms}
\usepackage[noend]{algpseudocode} 
\colorlet{linkequation}{blue}
\definecolor{darkgrayupdate}{gray}{0.25} 
\usepackage{footnote}
\makesavenoteenv{algorithm}
\makesavenoteenv{algorithmic}
\usepackage{diagbox} % For diagonal split in table cells
\usepackage{pifont} % For check and cross symbols
\usepackage{multirow}
\definecolor{darkgreen}{RGB}{0,128,0}
\usepackage[textsize=tiny]{todonotes}
\usepackage{booktabs}
\newcommand{\removed}[1]{}

\newcommand{\dLearn}{\Delta_{\mathsf{L}}}
\newcommand{\dRetain}{\Delta_{\mathsf{R}}}
\newcommand{\U}{\mathcal{U}}
\newcommand{\lr}{\mathtt{lr}}
\newcommand{\bs}{\mathtt{bs}}

\input{def}

\usepackage[suppress]{color-edits} % turn edits on   

\title{
Tune for Retention, Not Just Gain\\
SFT Forgets Far Less than You Think}

\author[1]{Nirmit Joshi\footnote{Correspondence: \texttt{nirmit@ttic.edu}.
Work done while interning at Apple. \url{https://github.com/nirmitj6/tune-for-retention-sft}}}
\author[1]{Nathan Srebro}
\author[2]{Eran Malach}

\affil[1]{\small 
Toyota Technological Institute at Chicago.}
\affil[2]{\small 
Apple.}

\date{}
\allowdisplaybreaks
\begin{document}
\maketitle
\vspace{-13mm}
\begin{abstract}
Catastrophic forgetting, the drastic loss of previously acquired capabilities when learning new ones, is a central challenge in language model fine-tuning. Supervised fine-tuning (SFT) is widely believed to be prone to such forgetting. Recent works have attributed this to the off-policy nature of SFT, arguing that on-policy learning is key to continual learning. In this paper, we challenge this premise. We show that while maximizing new-task performance can induce catastrophic forgetting, alternative hyperparameter choices (lower learning rate and batch size) can yield similar gains with little to no forgetting. With hyperparameters selected jointly for gain and retention, vanilla SFT retains prior capabilities far better than previously suggested and can be competitive with Self-Distillation Fine-Tuning (SDFT), an on-policy approach designed to mitigate forgetting. While SDFT can be more robust to hyperparameter choices, it also requires a retention-aware criterion to identify hyperparameter configurations that avoid catastrophic forgetting. Our results challenge the view that on-policy training is what enables learning with retention and raise broader questions about what drives catastrophic forgetting during fine-tuning.
\end{abstract}
\vspace{-5pt}

\begin{figure*}[!h]
    \centering
    \includegraphics[height=0.36\textwidth]
        {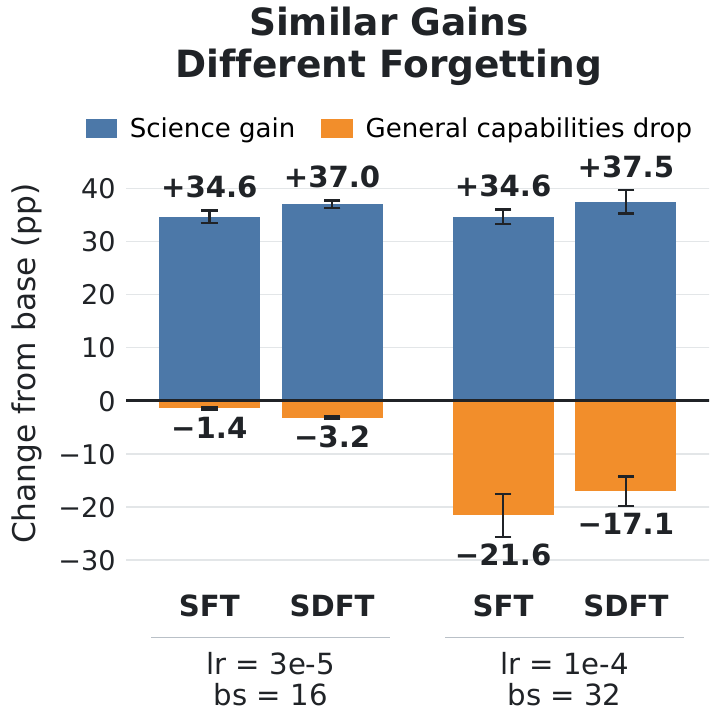}%
    \hspace{0.005\textwidth}%
    \includegraphics[height=0.36\textwidth]
        {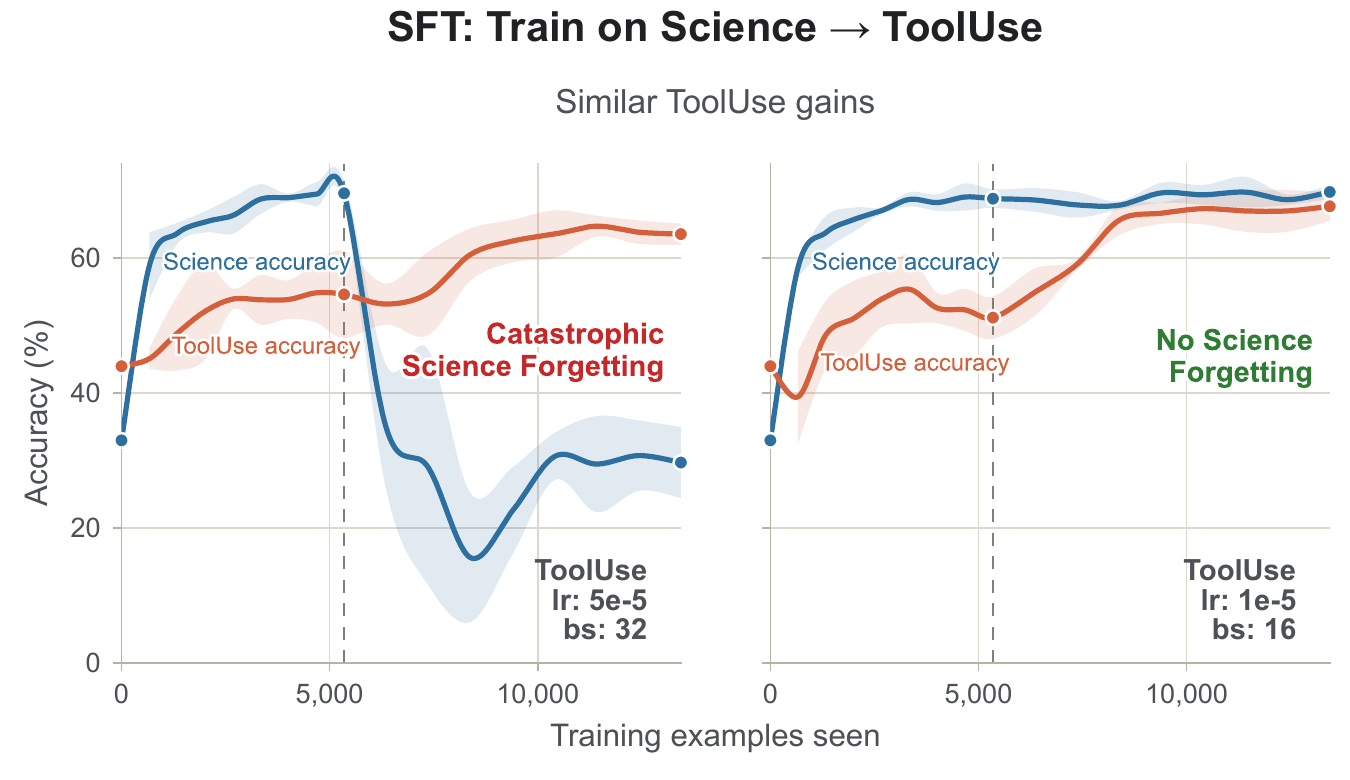}
    \caption{
    \textbf{Similar gains, different degrees of forgetting, from catastrophic to none.}
  (Left) For \emph{both} (off-policy) SFT and (on-policy) SDFT, Science fine-tuning with different hyperparameters yields similar
gains but markedly different retention of general capabilities. (Right) In sequential Science$\rightarrow$ToolUse training, adjusting only the ToolUse training hyperparameters allows off-policy SFT to learn ToolUse without forgetting previously learned Science. Lower learning-rate (and batch size) configurations achieve comparable gains with little or no forgetting.
}
\vspace{-9pt}
    \label{fig:intro-figure}
\end{figure*}

% It is widely believed that standard supervised fine-tuning (SFT) can cause catastrophic forgetting of a model’s prior capabilities. Several recent works have attributed this forgetting, at least in part, to the off-policy nature of SFT, and have proposed on-policy alternatives—such as reinforcement learning with GRPO when a verifier is available, or self-distillation when only demonstrations are available—as a way to learn new capabilities while better preserving old ones.
% In this work, we challenge this premise and ask two basic questions: Does SFT really forget? And is on-policy learning really what enables learning without forgetting? We show that, with appropriate hyperparameter selection, SFT can make substantial progress on new tasks while retaining much more of the model’s prior capabilities than commonly believed. Our results suggest that the apparent retention advantage of on-policy methods may not fundamentally arise from their on-policy nature, and motivate a re-examination of what actually drives forgetting and retention during continual fine-tuning.
%\nirmit{Other title suggestion: SFT Forgets Far Less Than You Think\\SFT  Retain Much More Than You Think\\
%Does SFT Really Forget? On-Policiness Is Not the Key to Continual Fine-Tuning}

\input{paper}
\newpage
\bibliographystyle{plainnat} 
\bibliography{bibliography.bib}
\appendix
% Use appendix names for the lettered headings, without changing main-text references.
\crefalias{section}{appendix}
\crefalias{subsection}{subappendix}
\clearpage
\section*{Appendix Table of Contents}
\startcontents[sections]
\printcontents[sections]{l}{1}{\setcounter{tocdepth}{2}}

\input{appendix}

%\clearpage
\end{document}

%% file: def.tex
\newtheorem*{theorem*}{Theorem}

\theoremstyle{definition}

\newtheoremstyle{myremark} % name
    {\topsep}                    % Space above
    {\topsep}                    % Space below
    {\it}                        % Body font
    {}                           % Indent amount
    {\bf}                        % Theorem head font
    {.}                          % Punctuation after theorem head
    {.5em}                       % Space after theorem head
    {}  % Theorem head spec (can be left empty, meaning normal)

\theoremstyle{definition}
\newtheorem{remark}{Remark}

\DeclareSymbolFont{rsfs}{U}{rsfs}{m}{n}
\DeclareSymbolFontAlphabet{\mathscrsfs}{rsfs}

%% file: paper.tex
\section{Introduction}\label{sec:intro}
Adapting language models to new domains, skills, and knowledge without forgetting their previously learned capabilities is a central challenge in fine-tuning. Designing methods for such adaptation is increasingly pressing as language models are continually updated to incorporate new capabilities, domain expertise, and knowledge. The simplest and most widely used adaptation approach is supervised fine-tuning (SFT), which is widely reported to cause \emph{``catastrophic forgetting''} of prior capabilities~\citep{luo2023empirical,li2024revisiting,ding2025improved,wu2025mitigating}. 

Recent work points to the \emph{off-policy} nature of SFT as a key culprit in catastrophic forgetting~\citep{shenfeld2026rlrazor,chen2026retaining}. Here, off-policy refers to the fact that SFT trains on demonstrations generated by an external expert or data-generating policy, rather than on outputs sampled from the model being fine-tuned. By contrast, \emph{on-policy} methods such as reinforcement learning (RL) train on trajectories sampled from the learner’s own current policy, so the training distribution evolves with the model itself. This has been argued to allow stronger retention~\citep{shenfeld2026rlrazor,chen2026retaining}. More recently, \citet{shenfeld2026self} brings this on-policy perspective to learning from demonstrations via self-distillation fine-tuning (SDFT). Using only a static set of teacher demonstrations, SDFT constructs a training signal from the student model’s own generations. \citeauthor{shenfeld2026self} demonstrates that this enables continual learning and avoids catastrophic forgetting. Taken together, these works put forward the following picture: off-policy SFT tends to forget, while on-policy training is a key mechanism for learning without forgetting~\citep{chen2026retaining,shenfeld2026self,shenfeld2026rlrazor}.

% Recent work points to ``off-policy'' nature of SFT as a key culprit~\citep{shenfeld2026rlrazor,chen2026retaining}. , contrasting it with the stronger retention of on-policy Reinforcement Learning (RL). Here, off-policy refers to the fact that SFT trains on demonstrations generated by an external expert or data-generating policy, rather than on outputs sampled from the model being fine-tuned. By contrast, on-policy methods such as RL train using the model’s own current generations, so the training distribution evolves together with the model. \cite{shenfeld2026self} bring this on-policy perspective to learning from demonstrations through self-distillation fine-tuning (SDFT)---this method uses a training set of static demonstrations from a teacher to construct an on-policy training signal from the student’s own generations, and show that this can enable ``continual learning'' and largely avoids catastrophic forgetting. 
 %This view is supported by empirical ablations, along with complementary heuristic and simplified theoretical arguments.

In this work, we directly challenge this emerging narrative. We ask: does SFT really catastrophically forget prior capabilities?
 %catorget as much as these works suggest? And is on-policiness truly what enables learning without forgetting? 

\paragraph{Results.} Our main findings suggest that SFT forgets far less than previously reported (\Cref{fig:intro-figure}). 
\begin{itemize}
\item  \textbf{Similar New-Task Gains, Different Forgetting.} In \Cref{sec:similar-gain-diff-forgetting}, we show that SFT configurations with comparable new-task gains can differ substantially in retention: lower learning rates (and batch sizes) cause only little forgetting, whereas aggressive ones can cause catastrophic forgetting. Insisting on maximizing new-task performance can trade substantial retention for only marginal additional gains.

% In \Cref{sec:similar-gain-diff-forgetting}, we show that SFT configurations with similar new-task gains can produce very different degrees of forgetting. In particular, the lower learning rate training can produce comparable gains with very little forgetting, but the aggressive ones can cause catastrophic forgetting. Insisting on maximizing new-task performance can trade away substantial retention for only a small additional gain.
\item \textbf{Comparison After Tuning for Both Gain and Retention.}
In \Cref{sec:main-results}, we introduce criteria that jointly use new-task gain and prior-capability retention to select checkpoints. Under such retention-aware tuning, SFT is highly competitive with SDFT for learning with retention. Although SDFT is more robust across hyperparameters, it still requires retention-aware selection: otherwise, it can also trade substantial retention for only marginal additional gains. 
\item \textbf{Cost-effective Retention-aware Tuning.}
In \Cref{sec:cost-effective-hp}, we present two practical strategies for hyperparameter selection that significantly reduce tuning costs without substantially compromising the quality of the selected checkpoints: validation on a small subset of prior capabilities and retention-guided hyperparameter search.

% \item \textbf{SFT forgets far less than previously suggested.} In \Cref{sec:RL-SDFT-GRPO}, we find that properly tuned vanilla SFT retains substantially more prior capability than existing comparisons suggest—often essentially all of it while still making strong new-task gains. This behavior persists across tasks and model families. As a result, the retention advantage of on-policy SDFT and GRPO over a carefully tuned SFT baseline is often absent, and otherwise far more humbling than the prevailing picture suggests.

 \item  
 \textbf{Implications.}
Our work calls for a systematic understanding of how hyperparameter
choices affect new-task learning and prior-capability retention in
continual fine-tuning. We outline key open questions in
\Cref{sec:discussion}. Our two main takeaways for continual
fine-tuning are:
\begin{tcolorbox}
\begin{itemize}
    \item[$\bullet$] \textbf{(Takeaway 1.)} Continual fine-tuning requires a different model-selection principle
    from standard fine-tuning: simply squeezing out gains on the new task
    can induce severe forgetting, for both SFT and SDFT.
     \item[$\bullet$] \textbf{(Takeaway 2.)} Our results challenge the view that on-policy learning is the
    mechanism that enables learning without forgetting: a fully
    off-policy method, vanilla SFT, can exhibit the same qualitative
    learning with retention, when hyperparameter-tuned for both retention and gain.
\end{itemize}
\end{tcolorbox}
% settings where such retention-aware selection may not be feasible, and raise some questions for future work. 
\end{itemize}
\section{Related Work}\label{sec:related-work}
A broad body of work reports forgetting after SFT
\citep{luo2023empirical,li2024revisiting,biderman2024lora}.
Several works propose variants of SFT to mitigate forgetting
\citep{li2024revisiting,huang2024mitigating,
sanyal2025upweighting,wu2025mitigating}.
Some studies report results under fixed training recipes
\citep{luo2023empirical,li2024revisiting,huang2024mitigating}
or select learning rates by new-task performance
\citep{sanyal2025upweighting}.
We study how methods should be compared under the sensitivity of
hyperparameter choices, assessing whether reported advantages
persist when each method is tuned under shared selection
criteria that jointly account for new-task gain and
prior-capability retention.

Prior work reports that learning and forgetting are sensitive
to learning-rate choices and proposes explanations for these
effects \citep{lin2026sft,rofin2026learning,marek2026forgetting}.
Most closely related, \citet{lin2026sft} show that smaller
learning rates can reduce general-capability degradation
while preserving comparable new-task performance.
Nevertheless, recent studies report substantial forgetting
with SFT and present on-policy methods as enabling strong
new-task learning with little forgetting
\citep{shenfeld2026rlrazor,chen2026retaining,shenfeld2026self}.
This raises the question of whether on-policy methods retain
prior capabilities across all hyperparameter configurations
that achieve strong new-task performance, and whether any
greater robustness provides a practical advantage over SFT.

% Recent work has suggested a particularly compelling explanation for this phenomenon: SFT is off-policy. The model is trained on trajectories generated by an external expert rather than on the outputs it would produce itself. In contrast, reinforcement-learning methods such as GRPO train on the model’s own generations and have been observed to acquire new capabilities while forgetting substantially less. Chen et al. [Retaining by Doing] argue that this use of on-policy data is precisely what underlies the retention advantage of RL over SFT. When explicit rewards or verifiers are unavailable, Shenfeld et al. propose self-distillation fine-tuning (SDFT), which similarly replaces off-policy demonstrations with on-policy supervision and reports substantially improved retention. Together, these results suggest an appealing principle for continual post-training: to learn without forgetting, train on-policy.

% In this section, we highlight several important effects in hyperparameter choices—and how the best hyperparameters are picked--that can materially change how continual fine-tuning methods are selected and compared. To make these effects directly comparable to prior results, we work in the same experimental settings considered by \citet{shenfeld2026self} and \citet{chen2026retaining}.
\section{Similar SFT Gains Can Exhibit Very Different Forgetting}\label{sec:similar-gain-diff-forgetting}

In this section, we show that hyperparameter configurations with similar new-task gains can nevertheless exhibit very different degrees of forgetting. We begin by discussing: 
% In this section, we show that hyperparameter configurations with similar new-task gains can nevertheless exhibit very different degrees of forgetting. This effect is large enough to change how baseline methods compare in the on-policy methods vs SFT gap in continual fine-tuning settings studied by \citet{shenfeld2026self} and \citet{chen2026retaining}. We revisit their experimental setups and use them to identify the effects that motivate careful evaluation for a  continual fine tuning study. \todo{As I mentioned, I don't like this framing of ``revisiting the baselines from Shenfeld''. It's good to borrow their setting and compare to their methods, but I think the message should be broader than this. Main finding is that if you pick the best hparams for the new task then indeed you get forgetting, but there are other choices that lead to significantly less forgetting with similar gains. Then, we should plot SFT with hparams tuned for new task vs SFT with hparams tuned for both, and not "SFT in Shenfeld" vs "our SFT".}

\paragraph{How Hyperparameter Choices Affect Gain and Retention?} The most important hyperparameter of SFT is the learning rate. At one extreme, a very small learning rate ($\lr$) may barely change the model: new-task gains are small, but forgetting on prior capabilities is also negligible, with $\lr=0$ as the limiting case of no learning and no forgetting. At the other extreme, a suitably large learning rate can gain in the adaptation task, but may also move the model far enough to damage prior capabilities. The interesting question lies in the intermediate regime: \emph{whether substantial new-task learning necessarily comes at the cost of substantial forgetting}. Note that batch size ($\bs$) interacts with the learning rate choice. At larger learning rates, larger batches can help stabilize the more aggressive updates. At smaller learning rates, smaller batches may suffice, and can even be preferable under a fixed epoch budget,\footnote{One can also vary the number of training epochs; we discuss this effect later. For the purpose of this discussion, we keep the epoch budget fixed, and throughout the paper this budget is two epochs.} because they allow the model to take enough small update steps to learn the new task.

 \paragraph{Similar Gains, Different Degrees of Forgetting.} We first show that these hyperparameter choices can yield similar new-task gains while differing substantially in how much prior capabilities the model forgets. We demonstrate this in the continual fine-tuning settings studied by \citet{shenfeld2026self} and \citet{chen2026retaining}, where the effect is large enough to change the apparent gap between SFT and on-policy methods.

 % Hyperparameter configurations of SFT that achieve comparable new-task performance can lie at very different points on the learning--retention tradeoff\todo{this sentence is hard to understand}. Some configurations nearly match, or even slightly exceed\todo{how can they exceed if we are tuning for new-task performance? is this because we are tuning on a validation set?}, the new-task performance of other high-performing configurations while retaining substantially more prior capability. Thus, similar new-task gains can correspond to very different degrees of forgetting on prior capabilities.

 \begin{figure}[!ht]
    \centering
    \begin{subfigure}[t]{0.495\linewidth}
        \centering
        \includegraphics[width=\linewidth]{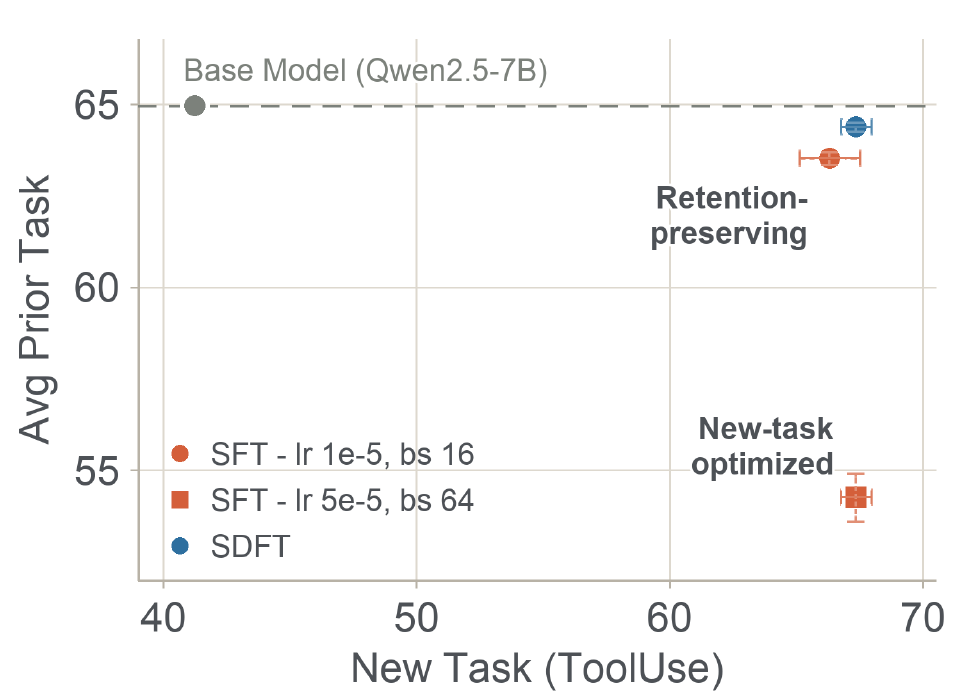}
        \caption{A setting from \cite{shenfeld2026self}.}
        \label{subfig:Shenfeld-ToolUse-Pareto}
    \end{subfigure}%
    \hspace{0.01\linewidth}%
    \begin{subfigure}[t]{0.495\linewidth}
        \centering
        \includegraphics[width=\linewidth]{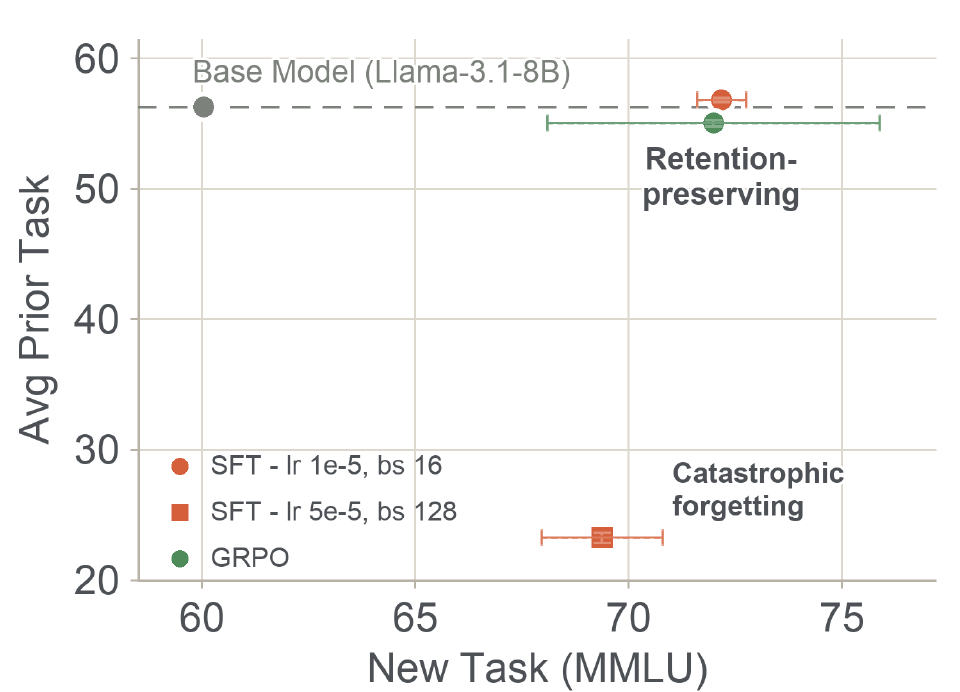}
        \caption{A setting from \cite{chen2026retaining}.}
        \label{subfig:chen-mmlu-8b}
    \end{subfigure}
    \caption{Similar new-task gains can correspond to substantially
    different degrees forgetting: demonstrated
    in the ToolUse-on-Qwen2.5-7B setting of
    \citet{shenfeld2026self} and the MMLU-on-Llama-3.1-8B setting of
    \citet{chen2026retaining}, with each point averaged over three seeds.}
    \label{fig:new-vs-old-tradeoff}
    \vspace{-3mm}
\end{figure}

 In \Cref{subfig:Shenfeld-ToolUse-Pareto}, we consider the setting of \citet{shenfeld2026self}, where adaptation is measured on ToolUse and retention is measured on a suite of prior capabilities; see \Cref{app:additional-baseline-revisits} for details. We include an SDFT run, within their hyperparameter sweep, that we believe reproduces their reported SDFT behavior (details in \Cref{app:revisit-shenfeld}). For SFT, two configurations achieve similar ToolUse performance but very different retention. The higher $\lr$ and $\bs$ configuration would be selected by maximizing ToolUse performance, but it exhibits catastrophic forgetting. The lower-$\lr$, lower-$\bs$ configuration is only 1.03 percentage points (pp) lower on ToolUse, but retains 9.27 pp more performance on the prior-capability suite.
  
 %This already shows that the apparent forgetting of SFT can change substantially across configurations that have similar performance on the new task.

We also see the same behavior in the setting of \citet{chen2026retaining}. In \Cref{subfig:chen-mmlu-8b}, we consider their strongest SFT-forgetting case, where Llama-3.1-8B is trained on MMLU as the new task and retention is evaluated on a separate suite of prior capabilities. Running SFT (with higher $\lr$ and $\bs$) and GRPO using the best provenance we could establish for their hyperparameter choices, we obtain learning and forgetting behavior qualitatively consistent with their report; see \Cref{rem:prov-chen-et-al.} in \Cref{app:additional-baseline-revisits}. However, a different SFT configuration achieves higher MMLU gains than their SFT baseline, matches GRPO’s new-task performance, and exhibits essentially no degradation of prior capabilities. \Cref{app:additional-baseline-revisits} further shows that the same pattern extends to additional SFT forgetting baselines from \citet{shenfeld2026self} and \citet{chen2026retaining}.

\begin{figure}[!t]
\centering
\captionsetup{skip=2pt}
\captionsetup[subfigure]{skip=1pt}

% Top row
\begin{subfigure}[t]{0.495\linewidth}
    \centering
    \includegraphics[width=\linewidth]{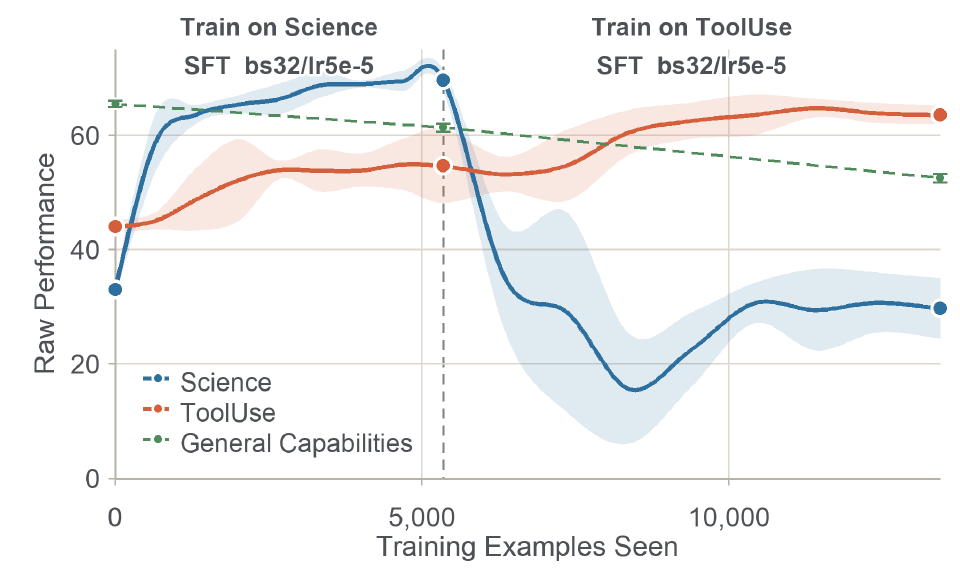}
    \caption{SFT in \cite{shenfeld2026self}.}
    \label{subfig:a}
\end{subfigure}%
\hspace{0.01\linewidth}%
\begin{subfigure}[t]{0.495\linewidth}
    \centering
    \includegraphics[width=\linewidth]{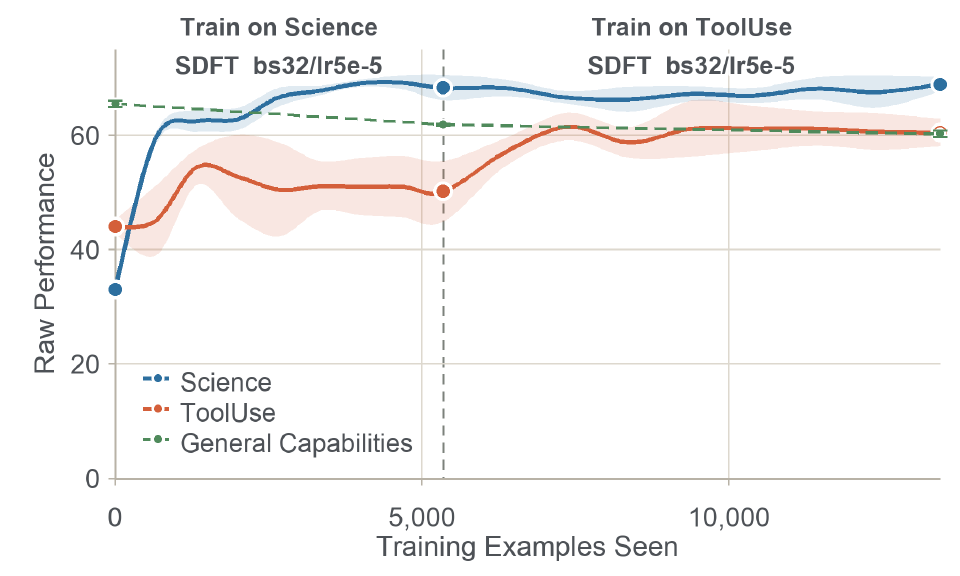}
    \caption{SDFT in \cite{shenfeld2026self}.}
    \label{subfig:b}
\end{subfigure}

\par\vspace{2pt}
\noindent
% Bottom row
\begin{minipage}[t]{0.495\linewidth}
    \vspace{0pt}
    \begin{subfigure}[t]{\linewidth}
        \centering
        \includegraphics[width=\linewidth]{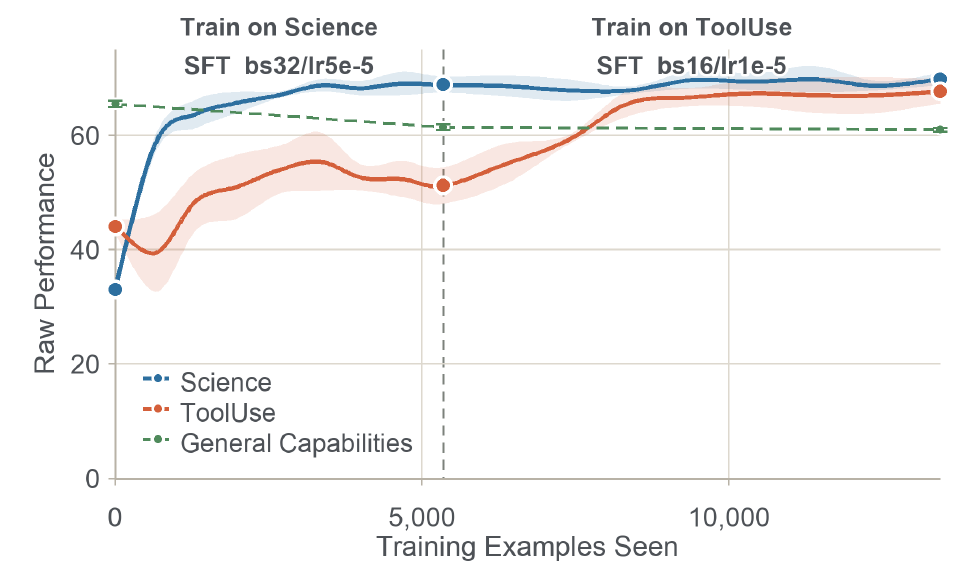}
        \caption{SFT with different hyperparameters.}
        \label{subfig:c}
    \end{subfigure}
\end{minipage}%
\hspace{0.01\linewidth}%
\begin{minipage}[t]{0.495\linewidth}
    \vspace{7pt}
    \captionsetup{skip=0pt}
    \caption{\textbf{Sequential fine-tuning on Science and ToolUse.}
    We report Science, ToolUse, and prior-capability performance,
    averaged over three seeds.
    \Cref{subfig:a,subfig:b} reproduce the qualitative SFT and SDFT
    behavior reported by \citet[Figure 3]{shenfeld2026self}.
    SFT with different ToolUse training hyperparameters
    (\Cref{subfig:c}) learns both tasks sequentially while retaining
    Science and general capabilities essentially as well as SDFT.}
    \label{fig:sequential-learning}
\end{minipage}
\end{figure}

\paragraph{Sequential Fine-Tuning Shows the Same Effect.}
We also consider the more challenging sequential-learning setup of \citet{shenfeld2026self}, where the model is fine-tuned on multiple tasks in succession (\Cref{fig:sequential-learning}). We follow their \citep[Figure 3]{shenfeld2026self} setting: training sequentially on Science and ToolUse, evaluating Science and ToolUse throughout, and evaluating the general prior-capability suite at the end of each phase.

\Cref{subfig:a,subfig:b} reproduce the qualitative SFT and SDFT behavior reported by \citet[Figure 3]{shenfeld2026self}.\footnote{We train on Science and ToolUse only, since the Medical data is unavailable. We report raw rather than normalized accuracy.} In \Cref{subfig:c}, we run SFT with lower $\lr$ and $\bs$ during ToolUse training. This SFT run learns both tasks sequentially while retaining performance on Science and the general prior-capability suite essentially as well as SDFT. By contrast, the SFT run in \Cref{subfig:a} would appear highly successful if judged only by its final ToolUse accuracy of 64.26\%, yet it suffers catastrophic forgetting on Science and degrades general capabilities.

\paragraph{Implications.} These examples show that continual fine-tuning comparisons can depend strongly hyperparameter choices. Unlike standard single-task fine-tuning, hyperparameter tuning should account for both new-task learning and retention of prior capabilities. These results also challenge the view that SFT forgets because it is intrinsically off-policy: in the settings above, much of the apparent forgetting is a property of the chosen SFT hyperparameters, not of SFT itself.

\paragraph{Our Questions.}
This raises several broader questions. How should continual fine-tuning methods be tuned and compared? Can methods other than SFT achieve their best new-task performance while still retaining prior capabilities, making them preferable? Or is there an inherent tradeoff between learning and retention? If such a tradeoff exists, how can we reliably select checkpoints based on validation performance on both new and prior tasks?

%Do some methods remain preferable because they achieve strong learning and retention more robustly across hyperparameter configurations?

\section{Our Experimental Setting}\label{sec:tuning-for-both}
In this section, we describe our main experimental setting.

\paragraph{New Skill-Acquisition Tasks.}
For new-task learning, we fine-tune on three qualitatively different tasks: Science, ToolUse, and Multistate Bar Examination (MBE). We choose tasks on which the base models are not already near ceiling performance, leaving room for meaningful new-task gains; otherwise, a method could appear to retain prior capabilities simply because it learns little. Science and ToolUse, adapted directly from \citet{shenfeld2026self}, are well suited for this purpose. As a third task, we include MBE questions to extend our evaluation to legal reasoning.

\begin{itemize}
    \item \textbf{Science.}
    We use the Chemistry Level-3 subset of SciKnowEval~\citep{feng2024sciknoweval}, following \citet{shenfeld2026self,shenfeld2026rlrazor}. The task consists of multiple-choice questions (MCQs).

    \item \textbf{ToolUse.}
    We use the dataset released by \citeauthor{shenfeld2026self}, a derivative of ToolAlpaca~\citep{tang2023toolalpaca}; see \Cref{rem:tooluse-dataset-gap}. The task requires selecting tool calls and their arguments.

    \item \textbf{Multistate Bar Examination (MBE).}
    We use the demonstration dataset of \citet{fernandes2025llamawalksbarefficient} for training and Bar Exam QA~\citep{zheng2025reasoning} for testing. The task consists of multiple-choice legal reasoning questions from the Multistate Bar Examination.
\end{itemize}

Science and MBE are evaluated by exact match on the answer option extracted from the model output. ToolUse is evaluated by matching the predicted tool calls and arguments to the released targets; see \Cref{sec:dataset-for-us} for details.

\paragraph{General Capability Retention Tasks.}
To measure retention, we use nine prior-capability categories spanning standard language-model abilities and safety behavior: (1) code generation (Code: MBPP and HumanEval), (2) instruction following (IFEval), (3) factuality (TruthfulQA), (4) general knowledge (MMLU), (5) commonsense reasoning (HellaSwag), (6) coreference reasoning (WinoGrande), (7) mathematical reasoning (MATH), (8) jailbreak robustness (WildJailbreak), and (9) harmful-prompt refusal (WildGuardTest). This suite combines the six prior-capability tasks used by \citet{shenfeld2026self} (1--6) with the three additional retention tasks used by \citet{chen2026retaining} (7--9). We modify only the Code category, pooling MBPP with HumanEval rather than using HumanEval alone to obtain a larger validation set for model selection.

Overall, our suite contains a good mix of tasks requiring structured outputs or free-form responses, such as code, instruction following, mathematics, and safety, and tasks evaluated using fixed answer options, such as MMLU, HellaSwag, WinoGrande, and TruthfulQA. This broad aggregate captures retention across diverse capabilities; we also report category-level results where relevant. Evaluation follows standard protocols; see \Cref{sec:dataset-for-us} for details.

\paragraph{Validation and Test Sets.}
For all benchmarks except ToolUse, we split the evaluation set into validation and test sets, using validation for hyperparameter and checkpoint selection and test for final reporting; see \Cref{tab:dataset-splits} for split sizes. ToolUse has only $97$ released evaluation examples, making further splitting too noisy, while validation examples drawn from the training distribution poorly represent its out-of-distribution evaluation set. We therefore use the entire evaluation set for both selection and reporting; a larger set of comparable difficulty would enable a reliable validation/test split, as demonstrated in other settings.\footnote{Sufficiently large validation sets are important for reliable selection. However, test-set overfitting appears less consequential for qualitative findings: using the full evaluation sets for both selection and reporting yields the same conclusions as selection based on sufficiently large validation sets.}

\paragraph{Algorithms.}
Our main goal is to compare SFT and SDFT. Given a demonstration $y^*$ for input $x$, SFT minimizes the next-token negative log-likelihood:
\begin{equation}\label{eq:SFT}
\mathcal{L}_{\mathrm{SFT}}(\theta)
=
-\sum_{t=1}^{|y^*|}
\log \pi_\theta\left(y_t^* \mid x, y_{<t}^*\right).
\end{equation}
This is an off-policy objective because the model is trained on prefixes from the demonstration trajectory $y^*$ rather than its own sampled trajectories. Thus, the training distribution need not match the distribution encountered by the model at generation time.

\citet{shenfeld2026self} propose self-distillation fine-tuning (SDFT), an on-policy fine-tuning algorithm. To understand SDFT, first suppose we had access to a strong teacher. After the student generates an answer $\hat{y} \sim \pi_\theta(\cdot \mid x)$, we could train it to imitate the teacher's next-token distribution along that generated trajectory:
\begin{equation}\label{eq:SDFT}
\mathcal{L}_{\mathrm{SDFT}}(\theta)
=
\mathbb{E}_{\hat{y} \sim \pi_\theta(\cdot \mid x)}
\left[
\sum_{t=1}^{|\hat{y}|}
\mathrm{KL}\left(
\pi_{\mathrm{T}}(\cdot \mid x, \hat{y}_{<t})
\,\middle\|\,
\pi_\theta(\cdot \mid x, \hat{y}_{<t})
\right)
\right].
\end{equation}
This is on-policy distillation \citep{agarwal2024policy} in the teacher--student setting. However, in our fine-tuning setting, we have only demonstrations, without access to such a teacher. \citet{shenfeld2026self} propose using a demonstration-conditioned version of the model as the teacher, a form of self-distillation. This teacher guides the student on the student's own generated prefixes. Specifically, the teacher uses an exponential moving average (EMA) of the student parameters, initialized from the base model and updated during student training:
\[
\pi_{\mathrm{T}}(\cdot \mid x, \hat{y}_{<t})
=
\pi_{\bar{\theta}}(\cdot \mid x, y^*, \hat{y}_{<t}),
\qquad
\bar{\theta}_{k}
=
(1-\alpha) \bar{\theta}_{k-1}
+
\alpha\theta_k,
\qquad
\bar{\theta}_0 = \theta_{\mathrm{base}},
\]
where $\alpha$ is the EMA hyperparameter. See also \Cref{rem:SDFT-obj-clrification} regarding the SDFT objective in \eqref{eq:SDFT}.

  % \item \textbf{GRPO (verifier, on-policy).}
    % For RL, the model samples completions from its current policy and is optimized against a
    % verifier-provided new-task reward. We use an on-policy policy-gradient objective, optionally
    % with a reference-policy regularizer:
    % \[
    % \mathcal{L}_{\mathrm{GRPO}}(\theta)
    % =
    % -\mathbb{E}_{\hat{y} \sim \pi_\theta(\cdot \mid x)}
    % \left[
    % \sum_t
    % A_t
    % \log \pi_\theta(\hat{y}_t \mid x,\hat{y}_{<t})
    % \right]
    % +
    % \beta\,
    % \mathrm{KL}\!\left(
    % \pi_\theta(\cdot \mid x)
    % \,\middle\|\,
    % \pi_{\mathrm{base}}(\cdot \mid x)
    % \right).
    % \]
    % Here $A_t$ is the GRPO advantage computed from rewards of multiple on-policy completions
    % sampled for the same prompt. In our setting, the reward is a verifier-based correctness signal,
    % e.g., final-answer or API-call correctness.
% \nirmit{This is very standard. I guess we can remove SFT and RL, and just say it is standard, and discuss SDFT as on-policy variant a bit more. Especially there was an error in Idan's paper, so clarifyinh and emphasizing what it is doing. Also, how is it on-policy and other variants like OPD and GKD.}\todo{you can leave the description of SFT and GRPO, depending on space}

\section{Main Results}
\label{sec:main-results}
In this section, we present our main experimental results. In \Cref{subsec:hp-slection}, we introduce hyperparameter selection rules that account for both new-task gain and prior-capability retention. We then compare SFT and SDFT under these criteria in \Cref{subsec:final-comparison}. Along the way, \Cref{subsec:lr-learning-forgetting} gives further evidence for the observation in \Cref{sec:similar-gain-diff-forgetting}, and how learning-rate choices shape both learning and forgetting. Additional factors and results at a larger model scale are discussed in \Cref{subsec:additional-obs}.

\subsection{Hyperparameter Selection for Gain and Retention}\label{subsec:hp-slection}
Let $\theta_{\mathrm{base}}$ denote the base model, and $\theta_{\mathrm{FT},A,\lambda}$ denote the model obtained by fine-tuning with
algorithm $A$ and hyperparameters $\lambda$. Let
$\mathrm{Acc}_{\mathrm{new}}(\theta)$ be accuracy on the new task, and let
$\mathrm{Acc}_{\mathrm{prior}}(\theta)$ be the average accuracy over the
prior-capability suite. We define
\begin{equation}\label{eq:learn-and-retention-changes}
\dLearn(A,\lambda)
:=
\mathrm{Acc}_{\mathrm{new}}(\theta_{\mathrm{FT},A,\lambda})
-
\mathrm{Acc}_{\mathrm{new}}(\theta_{\mathrm{base}}),
\qquad
\dRetain(A,\lambda)
:=
\mathrm{Acc}_{\mathrm{prior}}(\theta_{\mathrm{FT},A,\lambda})
-
\mathrm{Acc}_{\mathrm{prior}}(\theta_{\mathrm{base}}).
\end{equation}
Thus, $\dLearn$ (\emph{Learning} Delta) measures new-task gain, while $\dRetain$ (\emph{Retaining} Delta) measures forgetting. In particular, $-\dRetain$ is the net drop in the prior capabilities.

\paragraph{Joint Utility (Retention+Gain).} Our primary selection criterion is the joint learning--retention utility
\begin{equation}\label{eq:utility-LR}
\U_{\mathsf{LR}}(A,\lambda)
:=
\dLearn(A,\lambda)+\dRetain(A,\lambda).
\end{equation}
This criterion assigns equal weight to improving new-task accuracy and retaining
prior capabilities. Since the base-model terms are fixed, maximizing
$\U_{\mathsf{LR}}$ is equivalently maximizing
$\mathrm{Acc}_{\mathrm{new}}(\theta_{\mathrm{FT},A,\lambda})
+
\mathrm{Acc}_{\mathrm{prior}}(\theta_{\mathrm{FT},A,\lambda})$. For each algorithm, we estimate this utility on validation splits and select
\[
\lambda_A^\star
\in
\operatorname*{arg\,max}_{\lambda}
\widehat{\U}_{\mathsf{LR}}^{\mathrm{val}}(A,\lambda).
\]
We then evaluate $\theta_{\mathrm{FT},A,\lambda_A^\star}$ on the held-out test
sets and report its new-task accuracy, prior-capability accuracy, and
$\U_{\mathsf{LR}}^{\mathrm{test}}(A,\lambda_A^\star)$.

\paragraph{Maximizing Gains for a Forgetting Threshold.} 
Another criterion of practical interest is to allow a fixed small degradation in
prior capabilities, and subject to this constraint, maximize new-task gain.
For a tolerance $\epsilon \ge 0$, this corresponds to selecting
\begin{equation}\label{eq:constrained-selection}
\lambda_{A,\epsilon}^\star
\in
\operatorname*{arg\,max}_{\lambda}
\widehat{\dLearn}^{\mathrm{val}}(A,\lambda)
\quad
\text{s.t.}
\quad
-\widehat{\dRetain}^{\mathrm{val}}(A,\lambda) \leq \epsilon .
\end{equation}
This criterion is useful when an application specifies an acceptable forgetting
budget and asks for the largest possible improvement on the new task.

\subsection{Learning Rate Drives the Learning--Forgetting Tradeoff}
\label{subsec:lr-learning-forgetting}

\begin{figure}[t]
    \centering
    \includegraphics[width=\linewidth]{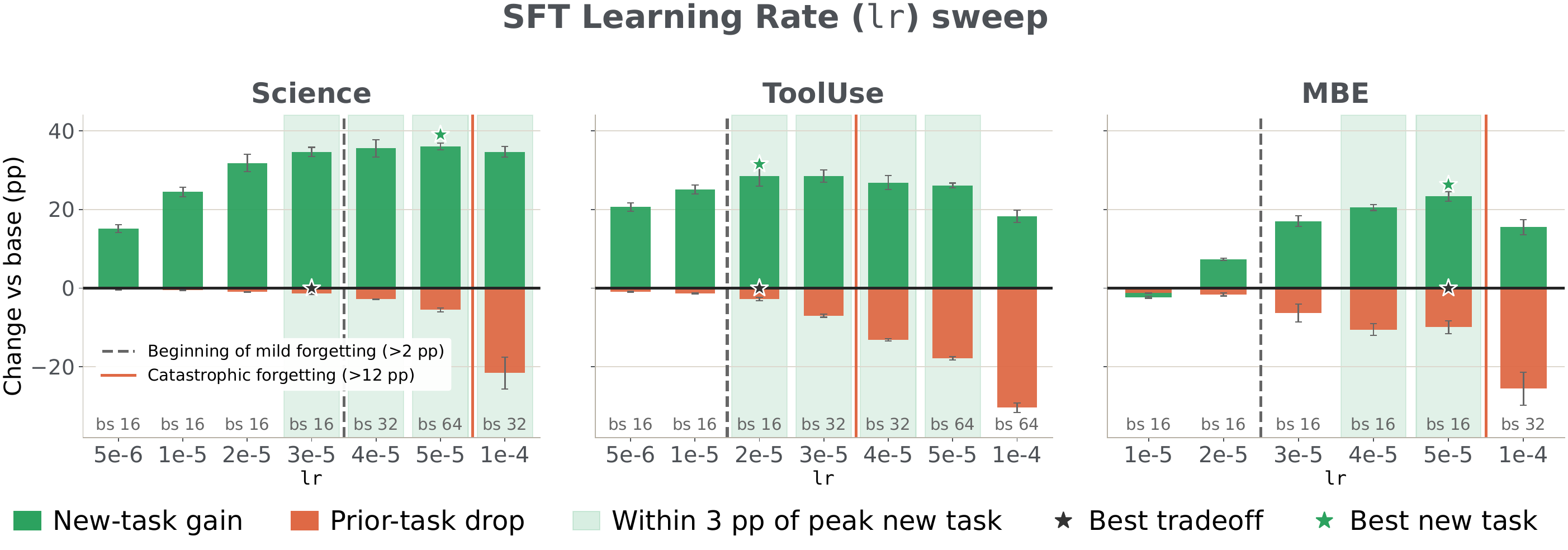}\\[3pt]
    \includegraphics[width=\linewidth]{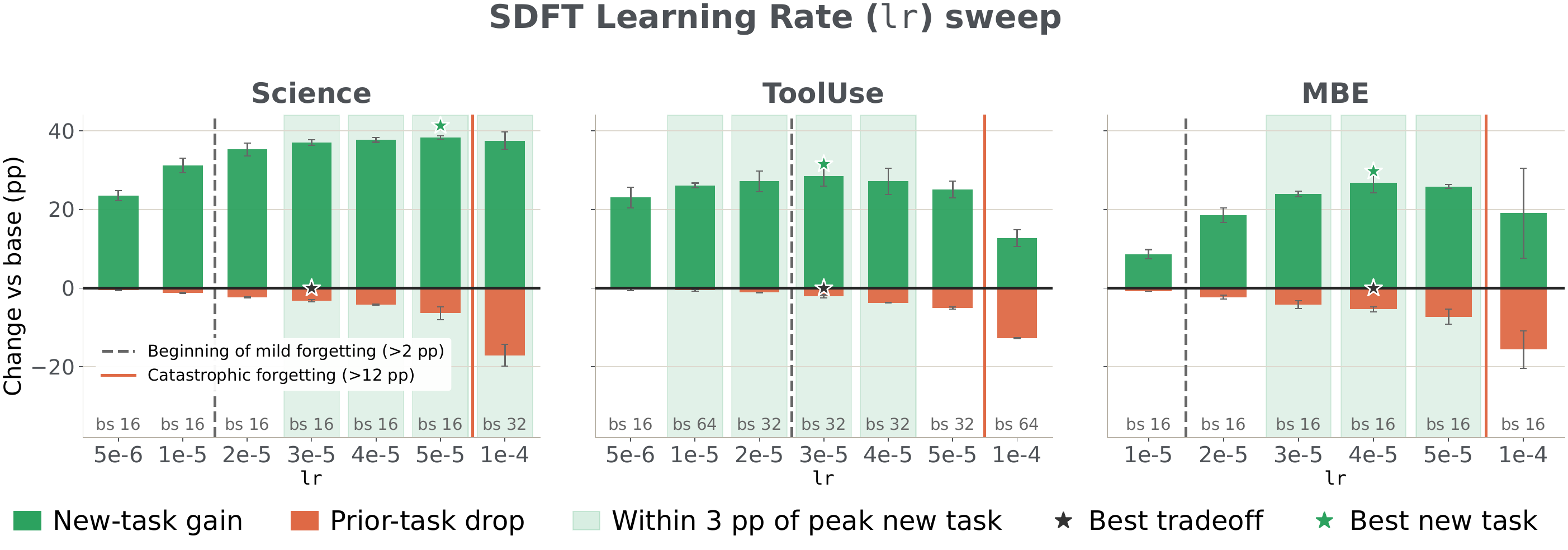}
    \captionsetup{skip=3pt}
    \caption{ $\lr$ sweep for Qwen2.5-7B:
    SFT (top) and SDFT (bottom).
    For each $\lr$, we select the $\bs$ that
    maximizes $\U_{\mathsf{LR}}$ from \eqref{eq:utility-LR};
    error bars show standard deviation across three reporting seeds.}
    \label{fig:qwen7b-lr-sweep-gain-drop}
\end{figure}

We further examine the role of learning rate in the gain--retention tradeoff discussed in \Cref{sec:similar-gain-diff-forgetting}. For each method, task, and learning rate, we fix training to $2$ epochs and select the best batch size and, for SDFT, EMA parameter $\alpha$ from \Cref{tab:hyperparameters} using $\U_{\mathsf{LR}}$.\footnote{Sweeping SDFT over $\alpha \in \{0.01,0.05\}$ gives it twice as many candidate runs as SFT, favoring SDFT in the selection budget.} \Cref{fig:qwen7b-lr-sweep-gain-drop} shows a consistent pattern for both methods: increasing $\lr{}$ improves new-task learning up to a point, but also increases forgetting of prior capabilities. This tradeoff has several consequences.
\begin{itemize}
    \item The best new-task checkpoint is \emph{not} always the best checkpoint
    for the joint utility criterion in \eqref{eq:utility-LR} (e.g., Science task
    for both SFT and SDFT).

    \item New-task-only selection can pick hyperparameters with catastrophic
    forgetting. The green shaded region marks learning rates whose new-task
    performance is within $3$ percentage points (pp) of the best observed
    new-task performance. We use this region to represent the range of
    hyperparameters that could plausibly be selected by cross-validation on the
    new task alone.\footnote{This is conservative: in the full sweep, other
    hyperparameters such as $\bs{}$ can also affect selection. Here, they are
    fixed for each $\lr{}$ by selecting the configuration with the highest joint
    utility, isolating the remaining ambiguity due to $\lr{}$ selection.}
    Within this region of nearly indistinguishable new-task performance,
    prior-task degradation can vary sharply and it also may overlap with the
    catastrophic-forgetting regime; see for example Science and ToolUse.

    \item SDFT achieves strong new-task performance, while retaining prior
    capabilities, \emph{more robustly} than SFT in the following sense:
    (i) SDFT has a broader range of learning rates between mild forgetting
    ($>2$ pp drop) and catastrophic forgetting ($>12$ pp drop), across all three
    tasks. This range can overlap more with the green shaded region
    (near-best new-task gain); and (ii) even beyond the
    catastrophic-forgetting threshold, SDFT suffers less severe degradation than
    SFT, especially on ToolUse and MBE.
\end{itemize}

\subsection{Comparing SFT vs SDFT after Tuning for Gain and Retention}
\label{subsec:final-comparison}

We now ask whether validation-based selection using the joint utility criterion in \eqref{eq:utility-LR} reliably identifies good hyperparameters. We compare this with selection based solely on new-task gain to quantify the cost of ignoring retention. In either case, we assess whether SDFT's greater hyperparameter robustness translates into sizable benefits over correctly tuned SFT.

\begin{figure}[!ht]
    \centering
    \includegraphics[width=\textwidth]{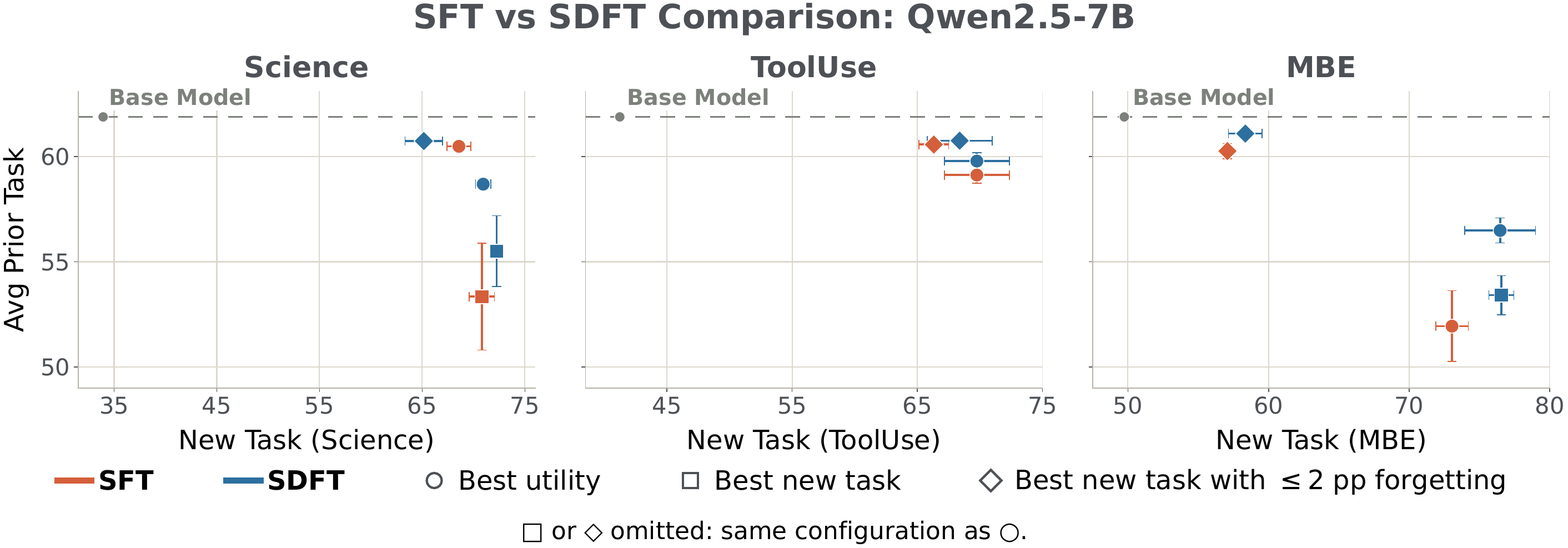}
    \caption{New-task versus average prior-task performance for Qwen2.5-7B. Each panel compares SFT and SDFT under three selection criteria: maximizing joint utility \eqref{eq:utility-LR}, maximizing new-task gain subject to a $2$ pp forgetting budget \eqref{eq:constrained-selection}, and maximizing new-task gain alone.}
    \label{fig:qwen7b-final-comparison}
\end{figure}

We evaluate SFT and SDFT over the full hyperparameter sweep in \Cref{tab:hyperparameters} under both selection criteria. \Cref{fig:qwen7b-final-comparison,tab:qwen7b-joint-vs-new-selection} present the results; we summarize the main observations below.

% what is the cost of not tuning Or does the m
% With the learning rate choice significantly determining the degree of forgetting for any learning level, in this sect
% The learning-rate sweeps above suggest that SDFT is less brittle than SFT: it often achieves high new-task gain over a wider range of hyperparameters, while suffering less prior-task degradation. We now ask whether this robustness translates into better selected checkpoints after tuning over the full hyperparameter sweep. In particular, we compare SFT and SDFT after selecting hyperparameters using the joint utility criterion in \eqref{eq:utility-LR}, which explicitly rewards new-task gain while penalizing prior-task degradation.

\begin{itemize}
    \item Under the joint-utility criterion, SFT is competitive with SDFT on Science and ToolUse, where the selected checkpoints exhibit only mild forgetting. In contrast, on MBE, where the best utility requires trading off new-task gain against prior-task forgetting, SDFT outperforms SFT.

    \item While selecting hyperparameters by new-task performance alone can coincide with the joint-utility choice, as in ToolUse, this is not always reliable and can hurt retention. On Science, new-task-only selection gives only a modest additional gain in $\dLearn$, but at a much larger retention cost in $-\dRetain$, especially in SFT. The same pattern also appears for SDFT on MBE, where new-task-only selection gains almost nothing in $\dLearn$ while losing several points in $-\dRetain$.
\end{itemize}

\begin{table}[t]
\centering
\small
\begin{tabular}{llccc|ccc|cc}
\toprule
& & \multicolumn{3}{c}{Joint utility} & \multicolumn{3}{c}{New-task only} & \multicolumn{2}{c}{New-only minus joint} \\
\cmidrule(lr){3-5}\cmidrule(lr){6-8}\cmidrule(lr){9-10}
Task & Method
& $\dLearn$ & $-\dRetain$ & $\mathcal{U}_{\mathsf{LR}}$
& $\dLearn$ & $-\dRetain$ & $\mathcal{U}_{\mathsf{LR}}$
& $\Delta\dLearn$ & $\Delta(-\dRetain)$ \\
\midrule
Science & SFT
& $34.6_{\pm 1.2}$ & $1.4_{\pm 0.3}$ & $\textbf{33.2}$
& $36.9_{\pm 0.9}$ & $8.5_{\pm 0.3}$ & $28.3$
& $2.2$ & $\textbf{7.1}$ \\
& SDFT
& $37.0_{\pm 0.7}$ & $3.2_{\pm 0.2}$ & $\textbf{33.8}$
& $38.3_{\pm 0.4}$ & $6.4_{\pm 1.7}$ & $31.9$
& $1.3$ & $\textbf{3.2}$ \\
\midrule
ToolUse & SFT
& $28.5_{\pm 2.6}$ & $2.8_{\pm 0.4}$ & $\textbf{25.8}$
& $28.5_{\pm 2.6}$ & $2.8_{\pm 0.4}$ & $25.8$
& $0.0$ & $0.0$ \\
& SDFT
& $28.5_{\pm 2.6}$ & $2.1_{\pm 0.4}$ & $\textbf{26.4}$
& $28.5_{\pm 2.6}$ & $2.1_{\pm 0.4}$ & $26.4$
& $0.0$ & $0.0$ \\
\midrule
MBE & SFT
& $23.3_{\pm 1.2}$ & $9.9_{\pm 1.7}$ & $\textbf{13.4}$
& $23.3_{\pm 1.2}$ & $9.9_{\pm 1.7}$ & $\textbf{13.4}$
& $0.0$ & $0.0$ \\
& SDFT
& $26.7_{\pm 2.5}$ & $5.4_{\pm 0.6}$ & $\textbf{21.3}$
& $26.8_{\pm 0.9}$ & $8.5_{\pm 0.9}$ & $\textbf{18.3}$
& $\textbf{0.1}$ & $\textbf{3.1}$ \\
\bottomrule
\end{tabular}
\caption{
Comparison of joint-utility selection and new-task-only selection. Subscripts denote standard deviation across three reporting seeds.
}
\label{tab:qwen7b-joint-vs-new-selection}
\end{table}

% \todo{if we lack space, I think this can go to the appendix and mentioned in a sentence in the main text} 

Even when maximizing new-task gain under a forgetting threshold, as defined in \eqref{eq:constrained-selection}, SDFT's advantage over SFT is limited to a narrower range of settings; see \Cref{tab:qwen7b-retention-budgets}. With a very small forgetting budget (2pp), both methods are restricted to low-forgetting configurations and often perform similarly within seed variability. Relaxing the budget (to 3pp) need not favor SDFT when SFT already achieves substantial new-task gains with little forgetting, as in Science and ToolUse. However, SDFT can have an advantage when further gains from SFT come with substantial forgetting, whereas SDFT achieves greater gains with only a small additional retention loss, as in MBE.

\begin{table}[!ht]
\centering
\small
\begin{tabular}{llcc}
\toprule
Task & Method
& $-\dRetain \le 2$ pp
& $-\dRetain \le 3$ pp \\
\midrule
Science & SFT
& $\mathbf{34.6}_{\pm 1.2}$
& $\mathbf{35.6}_{\pm 2.2}$ \\
& SDFT
& $\mathbf{31.2}_{\pm 1.8}$
& $\mathbf{36.0}_{\pm 0.5}$ \\
\midrule
ToolUse & SFT
& $\mathbf{25.1}_{\pm 1.2}$
& $\mathbf{28.5}_{\pm 2.6}$ \\
& SDFT
& $\mathbf{27.1}_{\pm 2.6}$
& $\mathbf{28.5}_{\pm 2.6}$ \\
\midrule
MBE & SFT
& $\mathbf{7.3}_{\pm 0.3}$
& $\mathbf{7.3}_{\pm 0.3}$ \\
& SDFT
& $\mathbf{8.6}_{\pm 1.2}$
& $\mathbf{18.5}_{\pm 1.9}$ \\
\bottomrule
\end{tabular}
\caption{Best new-task gain $\dLearn$ (pp) under fixed forgetting thresholds. Entries show means and standard deviations across three seeds.}
\label{tab:qwen7b-retention-budgets}
\vspace{-10pt}
\end{table}

\subsection{Additional Factors and Larger Model}\label{subsec:additional-obs}
We briefly discuss additional important observations.
\paragraph{Most of the drop is in generative capabilities.}
\Cref{fig:qwen7b-lr-sweep-drop-split} shows that most of the drop in prior
capabilities, including the catastrophic drop at larger learning rates, comes
from generative evaluations (IFEval, Code, Math, WildJailbreak,
WildGuardTest). In contrast, non-generative benchmarks (MMLU, HellaSwag,
WinoGrande, TruthfulQA) remain comparatively stable. This suggests that fine-tuning degrades the ability to produce task-valid generations more severely than performance on non-generative benchmarks.

\paragraph{Effect of the number of epochs.}
All our main experiments use $2$ epochs of training.\footnote{As illustrated in \Cref{subfig:a}, catastrophic forgetting
begins early in training. Overtraining is therefore not the primary issue, and simply reducing the number of epochs would not prevent
catastrophic forgetting.} We ask whether longer training at a lower learning rate can achieve
the same new-task gains without the additional forgetting caused by a higher learning rate. In \Cref{fig:qwen7b-science-2-vs-3-ep}, our runs suggest that training for $3$ epochs can marginally improve over $2$ epochs at the same learning rate. However, it still trails the gains achieved with $2$
epochs at larger learning rates. Thus, extending training from $2$ to $3$ epochs at lower learning rates does not fully recover the gains achieved at larger learning rates.

\paragraph{Effect of batch size.}
\Cref{fig:lr-bs-interaction} shows the interaction between learning rate
and batch size. Below the critical learning-rate threshold where
catastrophic degradation begins, we find that smaller batch sizes
generally match or improve new-task gains. We attribute this to the larger number of optimization steps (since the number of epochs is fixed). Importantly,
this does not appear to come at the cost of additional forgetting.

\paragraph{Effect of model size.}
Repeating our analysis on Qwen2.5-14B yields the same main findings:
catastrophic forgetting still occurs, but suitable hyperparameter
choices enable similar new-task gains with little or no forgetting;
see \Cref{app:14b}. This suggests that limited model capacity alone does not explain
catastrophic forgetting, and that increasing model size does not
eliminate the effects documented here or the importance of
hyperparameter selection.

\section{A Practical Recipe for Cost-Effective Hyperparameter Tuning}\label{sec:cost-effective-hp}

Our results highlight the importance of tuning hyperparameters \emph{jointly} for new-task \emph{gain} and prior-capability \emph{retention}. However, evaluating performance on such a large suite of general capabilities across many hyperparameter configurations can be prohibitively compute-intensive. Moreover, some capabilities we wish to preserve may even be implicit and lack explicit evaluation sets. How can we perform retention-aware hyperparameter tuning under these constraints?

The most promising possibility is to identify warning alarms during training on the new task that are completely independent of the prior-capability suite (e.g., based on training statistics). Such alarms could help identify and early-stop training for configurations likely to cause catastrophic forgetting. We tracked absolute and relative parameter drift using the $\ell_1$, $\ell_2$, and $\ell_\infty$ norms, as well as cross-entropy loss on the new task during training, but did not reveal noticeable differences between configurations that caused catastrophic forgetting and those that did not.

\begin{wrapfigure}{r}{0.35\textwidth}
    \centering
    \includegraphics[width=\linewidth]{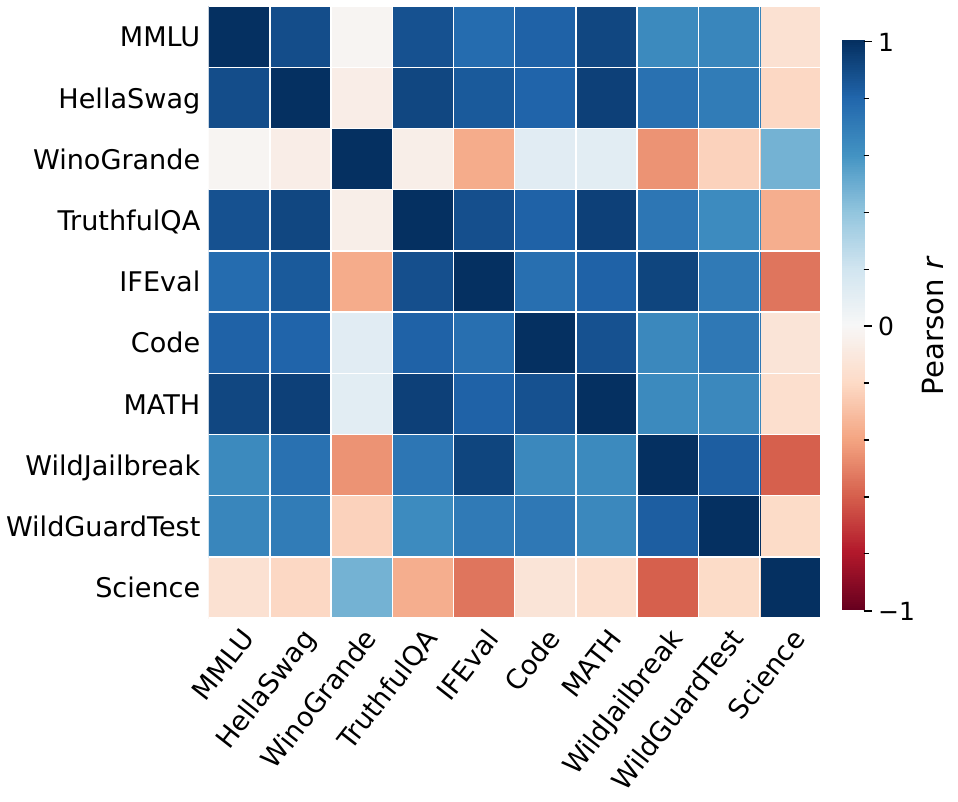}
        \vspace{-1cm}
\caption{Most prior capabilities exhibit strongly correlated performance
changes across SFT and SDFT configurations for Science.
See Fig \ref{fig:7b-correlations} for ToolUse and MBE.}
    \vspace{-1.2cm}
    \label{fig:science-capability-correlations}
\end{wrapfigure}
Identifying potential alarms during new-task training that are prior-task agnostic remains an important open question. For now, we present two cost-effective approaches that still use the prior-capability suite.
\subsection{Tuning on a Small Subset of Capabilities}
We first examine whether prior capabilities exhibit correlated patterns
of forgetting. \Cref{fig:science-capability-correlations} shows pairwise
Pearson correlations in performance changes from the base model (Qwen2.5-7B) across
configurations in the Science fine-tuning sweep of \Cref{tab:hyperparameters},
pooling SFT and SDFT runs. Most prior capabilities exhibit strongly
correlated accuracy changes, suggesting redundancy in the information
they provide for retention-aware tuning. This motivates using only a
small subset of them for validation and testing whether it recovers the hyperparameter choices made using the full prior-capability suite.

We use Science as a reference task to identify candidate validation subsets. For each of the $\binom{9}{3}=84$ triplets of prior capabilities, we repeat hyperparameter selection according to the utility criterion in \eqref{eq:utility-LR}, replacing the nine-category prior average with \emph{the average over the three selected capabilities}. We find that $23$ triplets recover the exact full-suite-selected checkpoint for each of SFT and SDFT on Science. From these, we fix $\{\text{MMLU}, \text{MATH}, \text{WildGuardTest}\}$ to span general knowledge, mathematical reasoning, and safety, as well as non-generative and generative evaluation. We then \emph{only} use this triplet for validation on ToolUse and MBE. It \emph{exactly} recovers the full-suite choices for both methods on both tasks, yielding zero utility regret.

We also note that this success extends beyond this particular choice: $12$ of the $23$ Science-qualified triplets of prior capabilities recover the full-suite choices for both methods on both ToolUse and MBE. For $20$ of the $23$ triplets, the worst-case utility regret across these two tasks and both methods is below $1$ pp; for the remaining three, it is below $1.4$ pp. These results provide strong empirical evidence that a small validation subset calibrated on one reference task can preserve full-suite selection decisions across other fine-tuning tasks. The same triplet also supports effective hyperparameter selection for the larger Qwen2.5-14B model (see \Cref{app:14b}).

\subsection{A Heuristic for Retention-Aware Hyperparameter Search}
Our next strategy seeks high-utility checkpoints by evaluating only a few configurations from the full sweep. Motivated by our observation that lower learning rates and smaller batch sizes can achieve strong gains with better retention, we propose the following protocol:
\begin{enumerate}
    \item Begin with the smallest batch size (e.g., $\bs=16$) and binary-search the ordered learning-rate grid using the criterion: If prior-capability drop exceeds $-\dRetain >2$ pp, search lower learning rates; otherwise, search higher learning rates. Continue until the search interval is empty.
    \item Select the evaluated configuration with the highest utility $\U_{\mathsf{LR}}$ in \eqref{eq:utility-LR}, then evaluate the next larger batch size at the same learning rate. If utility improves, repeat the learning-rate search at that batch size and consider the next larger batch size in the same way. Otherwise, stop.
\end{enumerate}
We apply this protocol independently to each task for SFT.\footnote{For SDFT, one can similarly fix the teacher EMA parameter, for example at $\alpha=0.01$, or repeat the protocol for each candidate $\alpha$.} In a retrospective evaluation on Qwen2.5-7B, the larger-batch check does not improve utility on any of the three tasks. The protocol therefore evaluates \emph{only four} configurations per task \emph{instead of $21$} in the full sweep (\Cref{tab:hyperparameters}), \emph{reducing} the number of evaluated configurations by $81.0\%$. Utility regret relative to full-sweep selection is zero on Science and ToolUse and $2.72$ pp on MBE.
\section{Discussion and Open Issues}\label{sec:discussion}

Our results show that SFT can learn new tasks while forgetting far less than previously suggested. Simple hyperparameter choices such as learning rate (and batch size) can produce very different degrees of forgetting at similar new-task gains, highlighting the importance of careful hyperparameter selection. We propose tuning hyperparameters jointly for new-task gain and prior-capability retention, rather than focusing primarily on squeezing out gains on the new task. Such retention-aware tuning appears necessary even for SDFT, making it an important principle for continual fine-tuning. Overall, our results shift the focus of continual fine-tuning from algorithm choice alone to the hyperparameter selection and evaluation. There are several important open questions that can be investigated further.
\paragraph{Whether and why SFT forgets.}
Independent studies have also reported hyperparameter effects
on forgetting (see \Cref{sec:related-work}), yet substantial
SFT forgetting remains a central finding in recent comparisons
with on-policy methods. Extending our comparisons to more
model families,\footnote{Our Llama-8B SFT results in
\Cref{fig:llama8b-lr-sweep-gain-drop} show the same pattern:
similar gains but different forgetting across learning rates
and batch sizes.} training settings, and low-rank fine-tuning
could clarify whether reported forgetting persists after
retention-aware tuning and when SFT can serve as a simple,
cost-effective strategy for continual adaptation. Our work does not attempt to explain why SFT forgets.
While prior studies have proposed some explanations
\citep{lin2026sft,rofin2026learning}, we believe a more
systematic understanding of why hyperparameter choices
affect learning and retention is needed.
Our results also suggest greater degradation on generative
evaluations. Non-generative evaluations of the same capabilities,
such as multiple-choice variants, could help distinguish
underlying knowledge loss from difficulty eliciting retained
knowledge.

\begin{figure}[!ht]
    \centering
    \includegraphics[scale=0.50]
    {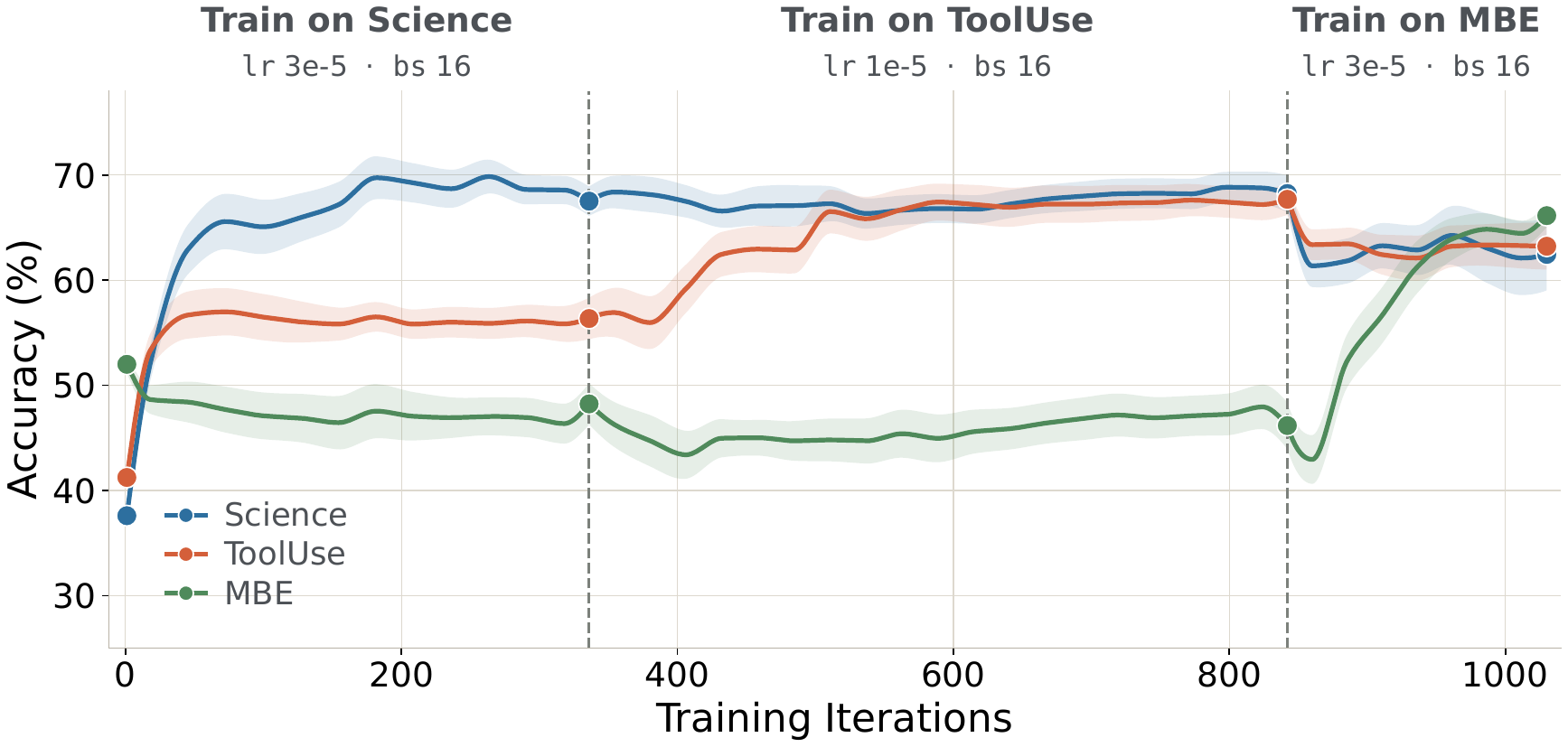}

    \caption{\textbf{Sequential SFT on Science$\rightarrow$ToolUse$\rightarrow$MBE.}
    We train Qwen2.5-7B for two epochs per task using the indicated
    hyperparameters. SFT learns each new task without catastrophic forgetting of what was previously learned.}
    \label{fig:sequential-sft-three-tasks}
\end{figure}

 \paragraph{Other algorithms in post-training.}
A central open question is whether an algorithm can avoid catastrophic
forgetting for all hyperparameter configurations that achieve strong
new-task performance. Such an algorithm would greatly reduce the
retention-aware tuning burden that remains for both SFT and SDFT.
Until then, understanding how hyperparameters affect learning and
retention remains essential. A related question is whether on-policy algorithms retain prior
capabilities more robustly across hyperparameters and whether
this robustness yields a meaningful practical advantage in learning
with retention over the highly cost-effective SFT. The same perspective
can also be applied to other post-training settings such as verifier-based RL, continual teacher--student distillation, preference
optimization such as DPO, and reward-model training.
\paragraph{Sequential fine-tuning.}
A more challenging setting is learning a sequence of tasks,
$A\rightarrow B\rightarrow C$, while retaining earlier tasks (and the
model's original capabilities). The main challenge is hyperparameter
selection across stages, both when tasks are known in advance and,
more challengingly, when they arrive online. We leave the development
of systematic hyperparameter selection strategies in both settings
to future work. In \Cref{fig:sequential-sft-three-tasks}, we provide
evidence that SFT can learn sequentially without catastrophic forgetting
and can serve as a strong baseline.

  % % Row B: 9, 100, 7, 11    (row total 127)
  % \node[cell, fill=blue!7]  at (1,-2) {9};
  % \node[cell, fill=blue!79] at (2,-2) {\textbf{100}};
  % \node[cell, fill=blue!6]  at (3,-2) {7};
  % \node[cell, fill=blue!9]  at (4,-2) {11};

  % % Row C: 10, 15, 91, 11   (row total 127)
  % \node[cell, fill=blue!8]  at (1,-3) {10};
  % \node[cell, fill=blue!12] at (2,-3) {15};
  % \node[cell, fill=blue!72] at (3,-3) {\textbf{91}};
  % \node[cell, fill=blue!9]  at (4,-3) {11};

  % % Row D: 6, 17, 24, 83    (row total 130)
  % \node[cell, fill=blue!5]  at (1,-4) {6};
  % \node[cell, fill=blue!13] at (2,-4) {17};
  % \node[cell, fill=blue!18] at (3,-4) {24};
  % \node[cell, fill=blue!64] at (4,-4) {\textbf{83}};
  % \end{tikzpicture}
  % \caption{Confusion matrix for the B200 (full-precision AdamW) run on the
  % science MCQ test set ($n=507$). Rows are gold answers, columns are model
  % predictions; cell intensity encodes count. Overall accuracy is $68.64\%$,
  % within $1.2$~pp of the 8-bit H100 run ($69.82\%$).}
  % \label{fig:cm-b200}
  % \end{figure}

%% file: appendix.tex
% \footnote{Although policy KL divergence on prior-task inputs provides another measure of behavioral change, it still requires access to those inputs, and larger divergence need not imply greater forgetting~\citep{chen2026retaining}. \nirmit{Revisit this.}} 

\section{Similar Gains, Different Forgetting (\texorpdfstring{\Cref{sec:similar-gain-diff-forgetting}}{Section 3}): Details}
\label{app:additional-baseline-revisits}
\suppressfloats[t]

We provide the experimental details for the settings of \citet{shenfeld2026self} and \citet{chen2026retaining} studied in \Cref{sec:similar-gain-diff-forgetting}.

\subsection{Settings of \texorpdfstring{\citet{shenfeld2026self}}{Shenfeld et al.}}
\label{app:revisit-shenfeld}

We follow the reported evaluation protocol of \citet{shenfeld2026self}, using Qwen2.5-7B-Instruct and their released Science and ToolUse data and graders. Retention averages accuracy on MMLU, HellaSwag, and WinoGrande; TruthfulQA-MC1; HumanEval \texttt{pass@1}; and IFEval instruction-level loose accuracy. Evaluation is zero-shot and greedy, without a chat template. Our base-model average of $64.97\%$ closely matches their $65.5\%$, which we take as evidence of reproducing their evaluation pipeline.

\begin{figure}[!ht]
    \centering
    \includegraphics[width=0.58\linewidth]{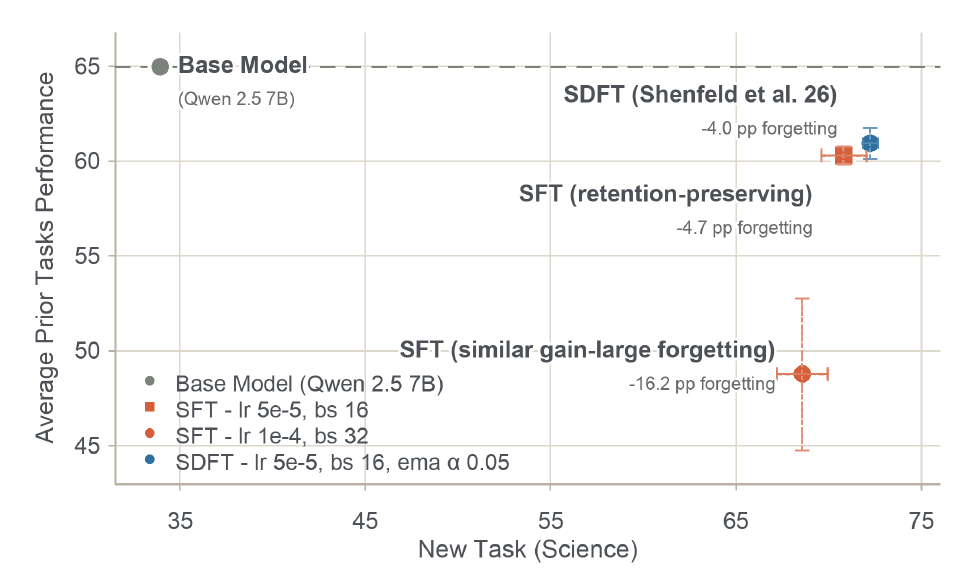}
    \caption{Science fine-tuning on Qwen2.5-7B (means over three seeds): SFT configurations with similar gains can have substantially different retention.}
    \label{fig:Shenfeld-Science-Pareto}
\end{figure}

\paragraph{Single-task fine-tuning.}
We follow the training setup of \citet{shenfeld2026self} and sweep hyperparameters to identify configurations close to their reported results.\footnote{Even when selecting solely for new-task gain, we had difficulty reproducing the reported extent of SFT forgetting. We do not know the source of this discrepancy. Nonetheless, such selection can trade prior capabilities for marginal new-task gains; see \Cref{subsec:final-comparison}.} We compare these with a higher-retention SFT run obtained by changing only the learning rate and batch size while maintaining similar new-task gains (\Cref{tab:revisit-shenfeld} and \Cref{subfig:Shenfeld-ToolUse-Pareto,fig:Shenfeld-Science-Pareto}).
\paragraph{Sequential fine-tuning.}
For \Cref{fig:sequential-learning}, we train on Science$\rightarrow$ToolUse for two epochs per task, restarting the schedule and warmup at each phase. We use learning rate $5{\cdot}10^{-5}$ following the authors' \href{https://github.com/idanshen/Self-Distillation/issues/9}{clarification}, with codebase defaults of batch size $32$ and SDFT EMA $\alpha=0.01$. In \Cref{subfig:c}, we change only the SFT ToolUse phase to learning rate $10^{-5}$ and batch size $16$.
\begin{table}[!htbp]
\centering
\small
\caption{ToolUse (\Cref{subfig:Shenfeld-ToolUse-Pareto}) and Science (\Cref{fig:Shenfeld-Science-Pareto}) fine-tuning on Qwen2.5-7B. All configurations are trained for 2 epochs; SDFT uses EMA $\alpha=0.01$ for ToolUse and $\alpha=0.05$ for Science. \textsc{Avg} is the unweighted mean over the six prior-capability benchmarks. Values are means over 3 seeds, with standard deviations shown for the new task and the average prior task.}
\label{tab:revisit-shenfeld}
\setlength{\tabcolsep}{4pt}

\begin{tabular}{lcccccccc}
\toprule
\multicolumn{9}{c}{\textbf{ToolUse}} \\
\midrule
Configuration & ToolUse & HSwag & HEval & IFEval & MMLU & TQA & WinoG & Avg \\
\midrule
Base (Qwen2.5-7B) & $41.24$ & $61.99$ & $64.02$ & $73.26$ & $71.76$ & $47.86$ & $70.96$ & $64.97$ \\
SDFT ($1{\cdot}10^{-5}$, bs $64$) & $67.35$ {\scriptsize $\pm0.59$} & $61.57$ & $65.44$ & $69.11$ & $71.57$ & $46.88$ & $71.77$ & $\mathbf{64.39}$ {\scriptsize $\pm0.12$} \\
SFT ($1{\cdot}10^{-5}$, bs $16$) & $66.32$ {\scriptsize $\pm1.19$} & $61.10$ & $61.18$ & $69.86$ & $71.57$ & $45.74$ & $71.77$ & $\mathbf{63.54}$ {\scriptsize $\pm0.17$} \\
SFT ($5{\cdot}10^{-5}$, bs $64$) & $67.35$ {\scriptsize $\pm0.59$} & $58.27$ & $47.36$ & $40.05$ & $70.48$ & $36.11$ & $73.32$ & $54.26$ {\scriptsize $\pm0.66$} \\
\bottomrule
\end{tabular}
\vspace{0.75em}
\begin{tabular}{lcccccccc}
\toprule
\multicolumn{9}{c}{\textbf{Science}} \\
\midrule
Configuration & Science & HSwag & HEval & IFEval & MMLU & TQA & WinoG & Avg \\
\midrule
Base (Qwen2.5-7B) & $33.93$ & $61.99$ & $64.02$ & $73.26$ & $71.76$ & $47.86$ & $70.96$ & $64.97$ \\
SDFT ($5{\cdot}10^{-5}$, bs $16$) & $72.25$ {\scriptsize $\pm0.41$} & $60.26$ & $70.53$ & $47.32$ & $69.97$ & $43.82$ & $73.80$ & $\mathbf{60.95}$ {\scriptsize $\pm0.81$} \\
SFT ($5{\cdot}10^{-5}$, bs $16$) & $70.81$ {\scriptsize $\pm1.23$} & $60.64$ & $66.87$ & $50.28$ & $70.27$ & $41.70$ & $72.19$ & $\mathbf{60.32}$ {\scriptsize $\pm0.47$} \\
SFT ($1{\cdot}10^{-4}$, bs $32$) & $68.57$ {\scriptsize $\pm1.38$} & $57.40$ & $47.97$ & $25.94$ & $52.80$ & $36.07$ & $72.53$ & $48.78$ {\scriptsize $\pm4.01$} \\
\bottomrule
\end{tabular}
\end{table}

\begin{remark}[SDFT objective]\label{rem:SDFT-obj-clrification}
\citet{shenfeld2026self} state the objective as
\[
\begin{aligned}
\mathcal{L}(\theta)
&= \mathrm{KL}\!\left(
\pi_\theta(\cdot\mid x)
\,\middle\|\,
\pi_{\mathrm{T}}(\cdot\mid x)
\right) = \mathbb{E}_{\hat{y}\sim\pi_\theta(\cdot\mid x)}
\left[
\sum_{t=1}^{|\hat{y}|}
\mathrm{KL}\!\left(
\pi_\theta(\cdot\mid x,\hat{y}_{<t})
\,\middle\|\,
\pi_{\mathrm{T}}(\cdot\mid x,\hat{y}_{<t})
\right)
\right].
\end{aligned}
\]
The authors later \href{https://github.com/idanshen/Self-Distillation#updates}{clarified} that their experiments sample student trajectories on-policy but compute the per-token forward KL from teacher to student, as in \eqref{eq:SDFT}.
\end{remark}

\begin{figure}[t]
\centering
\captionsetup{skip=2pt}
\captionsetup[subfigure]{skip=1pt}
\begin{subfigure}[t]{0.48\linewidth}
    \centering
    \includegraphics[width=\linewidth]{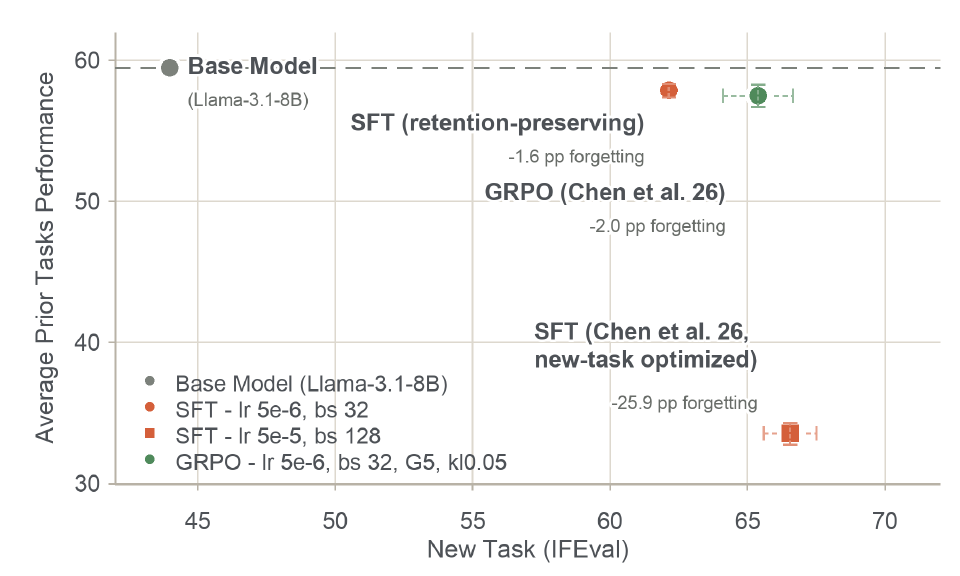}
    \caption{IFEval, Llama-3.1-8B.}
    \label{subfig:chen-ifeval-8b}
\end{subfigure}
\hfill
\begin{subfigure}[t]{0.48\linewidth}
    \centering
    \includegraphics[width=\linewidth]{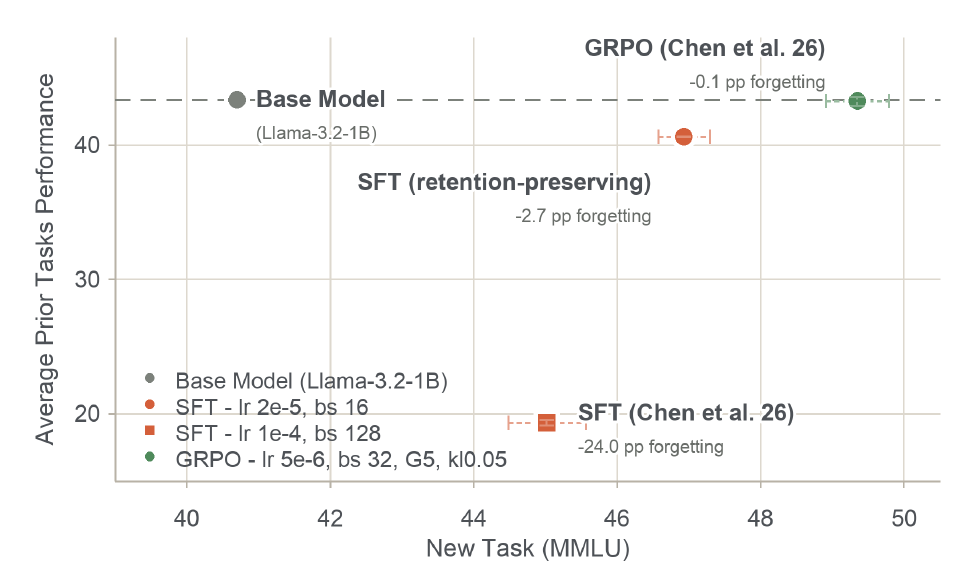}
    \caption{MMLU, Llama-3.2-1B.}
    \label{subfig:chen-mmlu-1b}
\end{subfigure}
\par\vspace{3pt}
\begin{minipage}[t]{0.48\linewidth}
    \vspace{0pt}
    \begin{subfigure}[t]{\linewidth}
        \centering
        \includegraphics[width=\linewidth]{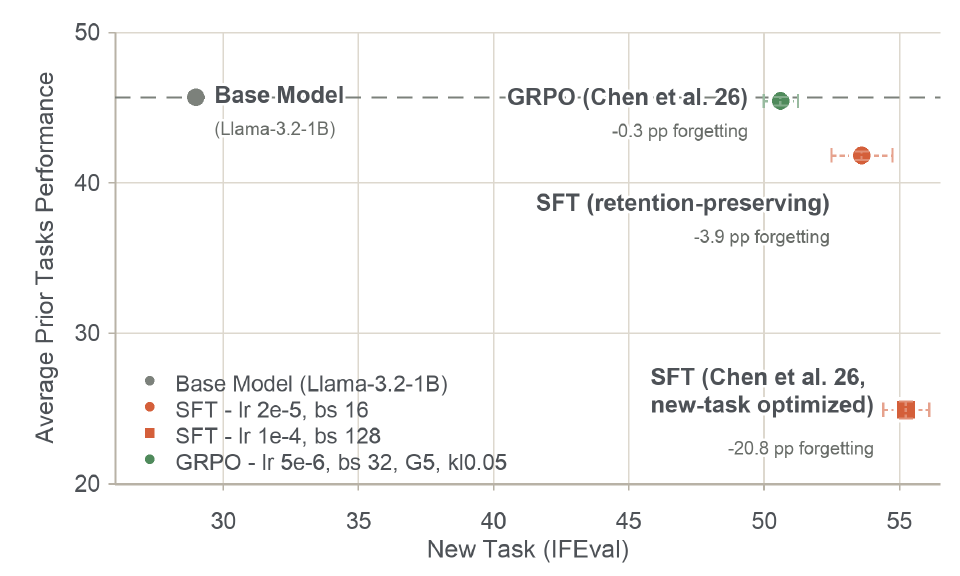}
        \caption{IFEval, Llama-3.2-1B.}
        \label{subfig:chen-ifeval-1b}
    \end{subfigure}
\end{minipage}
\hfill
\begin{minipage}[t]{0.48\linewidth}
    \vspace{7pt}
    \caption{\textbf{Additional comparisons in the settings of \citet{chen2026retaining}.} Each point is a mean over three seeds. Configurations and per-task results are given in \Cref{tab:chen-revisit}.}
    \label{fig:chen-additional-comparisons}
\end{minipage}
\end{figure}

\subsection{Settings of \texorpdfstring{\citet{chen2026retaining}}{Chen et al.}}
\label{subsec:revisitng-chen}

We revisit MMLU and IFEval fine-tuning on Llama-3.1-8B-Instruct and Llama-3.2-1B-Instruct.\footnote{We omit Countdown fine-tuning because the released SFT demonstrations do not contain reasoning traces.} We use the released training data and evaluation code of \citet{chen2026retaining}, including their WildGuard judge. Retention is the unweighted mean over the five non-target tasks among MMLU, IFEval, Countdown, MATH, WildJailbreak, and WildGuardTest.

\paragraph{Training configurations.}
SFT uses AdamW for two epochs with a cosine learning-rate schedule, warmup ratio $0.03$, gradient clipping $1.0$, and maximum sequence length $4096$. The comparison configurations use batch size $128$ and learning rates $5{\cdot}10^{-5}$ for 8B and $10^{-4}$ for 1B. Our higher-retention SFT configurations change only learning rate and batch size (\Cref{tab:chen-revisit}). GRPO uses two epochs, a constant learning rate of $5{\cdot}10^{-6}$, batch size $32$, $G=5$ generations per prompt, temperature $0.8$, and a $1024$-token generation budget. Its KL coefficient is $0$ for 8B MMLU and $0.05$ otherwise.

\begin{remark}[Comparison configurations]\label{rem:prov-chen-et-al.}
The SFT comparison learning rates follow the authors' \href{https://github.com/princeton-pli/retaining-by-doing/blob/main/scripts/sft.sh}{released training script}. We report our own three-seed reruns, which reproduce the qualitative pattern of substantial forgetting for these SFT configurations and stronger retention for GRPO.
\end{remark}

\Cref{tab:chen-revisit} reports all four settings, with corresponding learning--retention comparisons in \Cref{subfig:chen-mmlu-8b,fig:chen-additional-comparisons}. We discuss the results below.
\paragraph{8B MMLU (\Cref{subfig:chen-mmlu-8b}).}
Our higher-retention SFT configuration improves both learning and retention: new-task gain increases from $9.34$ to $12.13$ pp, while a $33.00$ pp prior-capability drop becomes a $0.55$ pp improvement. It matches GRPO's $11.95$ pp gain while preserving prior capabilities, making SFT competitive in this setting through hyperparameter changes alone.

\paragraph{8B IFEval (\Cref{subfig:chen-ifeval-8b}).}
The comparison SFT configuration achieves the largest IFEval gain ($22.56$ pp), but incurs a $25.92$ pp prior-capability drop. Our higher-retention configuration achieves an $18.16$ pp gain with only a $1.60$ pp drop. Its retention is comparable to GRPO's $1.97$ pp drop, although GRPO achieves a larger new-task gain of $21.39$ pp.

\paragraph{1B MMLU (\Cref{subfig:chen-mmlu-1b}).}
Our higher-retention SFT configuration increases new-task gain from $4.32$ to $6.23$ pp while reducing the prior-capability drop from $24.00$ to $2.72$ pp. Hyperparameter changes substantially narrow the SFT--GRPO gap, although GRPO remains stronger here, achieving an $8.65$ pp gain with a $0.09$ pp drop.

\paragraph{1B IFEval (\Cref{subfig:chen-ifeval-1b}).}
Our higher-retention SFT configuration preserves most of the new-task gain ($24.61$ versus $26.25$ pp) while reducing the prior-capability drop from $20.77$ to $3.88$ pp. It achieves greater new-task gain than GRPO ($21.60$ pp), while GRPO retains prior capabilities better, with a $0.26$ pp drop. The comparison therefore exposes a learning--retention tradeoff between the selected configurations.

\begin{table}[!ht]
\centering
\small
\caption{Results for MMLU and IFEval settings of \citet{chen2026retaining}. ``Prior'' is the average over the five non-target tasks. $\Delta$ rows report signed changes relative to the corresponding base model: positive and negative values indicate the gains and drops respectively.}
\label{tab:chen-revisit}
\setlength{\tabcolsep}{2.5pt}

\begin{subtable}[t]{0.48\textwidth}
\centering
\caption{MMLU, Llama 3.1 8B}
\begin{tabular}{llcc}
\toprule
Run & & MMLU & Prior \\
\midrule
Base & score & $60.04$ & $56.27$ \\
\midrule
SFT, $1{\cdot}10^{-5}$/bs16
& score & $72.17_{\scriptscriptstyle\pm0.57}$ & $56.82_{\scriptscriptstyle\pm0.17}$ \\
& $\Delta$ & $+12.13_{\scriptscriptstyle\pm0.57}$ & $+0.55_{\scriptscriptstyle\pm0.17}$ \\
SFT, $5{\cdot}10^{-5}$/bs128
& score & $69.38_{\scriptscriptstyle\pm1.42}$ & $23.27_{\scriptscriptstyle\pm0.39}$ \\
& $\Delta$ & $+9.34_{\scriptscriptstyle\pm1.42}$ & $-33.00_{\scriptscriptstyle\pm0.39}$ \\
GRPO
& score & $71.99_{\scriptscriptstyle\pm3.90}$ & $55.05_{\scriptscriptstyle\pm0.23}$ \\
& $\Delta$ & $+11.95_{\scriptscriptstyle\pm3.90}$ & $-1.22_{\scriptscriptstyle\pm0.23}$ \\
\bottomrule
\end{tabular}
\end{subtable}
\hfill
\begin{subtable}[t]{0.48\textwidth}
\centering
\caption{IFEval, Llama 3.1 8B}
\begin{tabular}{llcc}
\toprule
Run & & IFEval & Prior \\
\midrule
Base & score & $43.99$ & $59.48$ \\
\midrule
SFT, $5{\cdot}10^{-6}$/bs32
& score & $62.15_{\scriptscriptstyle\pm0.28}$ & $57.88_{\scriptscriptstyle\pm0.47}$ \\
& $\Delta$ & $+18.16_{\scriptscriptstyle\pm0.28}$ & $-1.60_{\scriptscriptstyle\pm0.47}$ \\
SFT, $5{\cdot}10^{-5}$/bs128
& score & $66.55_{\scriptscriptstyle\pm0.96}$ & $33.56_{\scriptscriptstyle\pm0.76}$ \\
& $\Delta$ & $+22.56_{\scriptscriptstyle\pm0.96}$ & $-25.92_{\scriptscriptstyle\pm0.76}$ \\
GRPO
& score & $65.38_{\scriptscriptstyle\pm1.28}$ & $57.51_{\scriptscriptstyle\pm0.78}$ \\
& $\Delta$ & $+21.39_{\scriptscriptstyle\pm1.28}$ & $-1.97_{\scriptscriptstyle\pm0.78}$ \\
\bottomrule
\end{tabular}
\end{subtable}

\vspace{2mm}

\begin{subtable}[t]{0.48\textwidth}
\centering
\caption{MMLU, Llama 3.2 1B}
\begin{tabular}{llcc}
\toprule
Run & & MMLU & Prior \\
\midrule
Base & score & $40.70$ & $43.36$ \\
\midrule
SFT, $2{\cdot}10^{-5}$/bs16
& score & $46.93_{\scriptscriptstyle\pm0.36}$ & $40.64_{\scriptscriptstyle\pm0.02}$ \\
& $\Delta$ & $+6.23_{\scriptscriptstyle\pm0.36}$ & $-2.72_{\scriptscriptstyle\pm0.02}$ \\
SFT, $1{\cdot}10^{-4}$/bs128
& score & $45.02_{\scriptscriptstyle\pm0.54}$ & $19.36_{\scriptscriptstyle\pm0.21}$ \\
& $\Delta$ & $+4.32_{\scriptscriptstyle\pm0.54}$ & $-24.00_{\scriptscriptstyle\pm0.21}$ \\
GRPO
& score & $49.35_{\scriptscriptstyle\pm0.44}$ & $43.27_{\scriptscriptstyle\pm0.35}$ \\
& $\Delta$ & $+8.65_{\scriptscriptstyle\pm0.44}$ & $-0.09_{\scriptscriptstyle\pm0.35}$ \\
\bottomrule
\end{tabular}
\end{subtable}
\hfill
\begin{subtable}[t]{0.48\textwidth}
\centering
\caption{IFEval, Llama 3.2 1B}
\begin{tabular}{llcc}
\toprule
Run & & IFEval & Prior \\
\midrule
Base & score & $29.00$ & $45.69$ \\
\midrule
SFT, $2{\cdot}10^{-5}$/bs16
& score & $53.61_{\scriptscriptstyle\pm1.12}$ & $41.81_{\scriptscriptstyle\pm0.28}$ \\
& $\Delta$ & $+24.61_{\scriptscriptstyle\pm1.12}$ & $-3.88_{\scriptscriptstyle\pm0.28}$ \\
SFT, $1{\cdot}10^{-4}$/bs128
& score & $55.25_{\scriptscriptstyle\pm0.85}$ & $24.92_{\scriptscriptstyle\pm0.58}$ \\
& $\Delta$ & $+26.25_{\scriptscriptstyle\pm0.85}$ & $-20.77_{\scriptscriptstyle\pm0.58}$ \\
GRPO
& score & $50.60_{\scriptscriptstyle\pm0.63}$ & $45.43_{\scriptscriptstyle\pm0.29}$ \\
& $\Delta$ & $+21.60_{\scriptscriptstyle\pm0.63}$ & $-0.26_{\scriptscriptstyle\pm0.29}$ \\
\bottomrule
\end{tabular}
\end{subtable}
\end{table}

\section{Additional Details for Our Experiments\texorpdfstring{ (\Cref{sec:tuning-for-both})}{}}\label{sec:additional-details-for-exp}

We now describe dataset details in \Cref{sec:dataset-for-us} and hyperparameter choices in \Cref{subsec:hp-choices} for our experiments.
 
 \subsection{Dataset Curation}\label{sec:dataset-for-us}

   \paragraph{Science.}
We use the Chemistry L3 split of SciKnowEval~\citep{feng2024sciknoweval}, following \citet{shenfeld2026self}. Each example is a four-option question, and the model is scored by exact match after emitting the answer letter inside an \texttt{<answer>}\dots\texttt{</answer>} field. We use the released $2{,}674$ training examples and split the $507$ held-out examples into validation and test sets of sizes $100$ and $407$, respectively, as shown in \Cref{tab:dataset-splits}.

\paragraph{ToolUse.}
We use the ToolAlpaca-derived dataset released by \citet{shenfeld2026self}, so that our results are directly comparable to theirs. The released training set contains $4{,}046$ examples, and the released evaluation set contains $97$ examples. Outputs are parsed into tool actions and JSON action inputs and scored by exact match against the gold actions and inputs. Because the released ToolUse evaluation set is small, we do not split it into separate validation and test subsets. We also found that validation sets drawn from the ToolUse training distribution are out-of-distribution relative to the released evaluation set and do not reliably select good checkpoints. Therefore, for this task only, we select checkpoints using the same released evaluation set on which we report performance.

\begin{remark}[ToolUse evaluation set]\label{rem:tooluse-dataset-gap}
The released ToolUse evaluation set differs from ToolAlpaca's original \texttt{eval\_simulated.json}. Of the $97$ released evaluation examples, $68$ are drawn verbatim from the original ToolAlpaca evaluation set, while $29$ do not appear in that set. Conversely, $32$ of ToolAlpaca's original $100$ simulated evaluation examples are omitted. Many omitted examples involve multi-step or dependent tool calls and remain difficult across the methods we study, while the added examples are substantially easier to improve on under fine-tuning. All methods in our experiments use the same released evaluation set. Absolute ToolUse accuracies should be interpreted in light of its construction.
\end{remark}

\paragraph{Multistate Bar Examination (MBE).}
Our training set consists of $1{,}501$ examples with distilled explanations from \texttt{HolySaint/MBE-exam-questions}~\citep{fernandes2025llamawalksbarefficient}. The evaluation set comes from the \texttt{qa} configuration of Bar Exam QA~\citep{zheng2025reasoning}, released as \texttt{reglab/barexam\_qa}. After removing $44$ exact duplicates of training items from $1{,}194$ usable evaluation questions, we split the remaining $1{,}150$ into validation ($230$) and test ($920$) sets (\Cref{tab:dataset-splits}). Many remaining questions are rewordings of training items, so MBE accuracy reflects a mixture of recall and generalization. Learning domain-specific legal knowledge and applying it to closely related questions are meaningful adaptation objectives. Therefore, MBE provides a valid setting for measuring how such adaptation affects retention of prior capabilities. Outputs are scored using the same answer-tag exact-match protocol as Science.

  % \paragraph{Math (AMC/AIME).}
  % The math task is drawn from NuminaMath-CoT~\citep{numina_math_datasets}, restricting to the
  % \texttt{amc\_aime} source (AMC/AIME competition problems paired with worked
  % chain-of-thought solutions). We extract each problem's final answer from its
  % \verb|\boxed{}| expression and \emph{retain only problems whose answer is
  % numeric} (integer, fraction, or decimal), for which exact-match / symbolic
  % equivalence grading is reliable; problems with multiple-choice-letter,
  % free-form, or symbolic-expression answers are discarded, as they cannot be
  % graded automatically without ambiguity. Crucially, we apply the \emph{same}
  % numeric filter to the \emph{training} set, not just the evaluation set. This is
  % because SFT imitates the surface form of its demonstrations: training on answer
  % formats that never appear at evaluation would unfairly handicap SFT, so matching
  % the train and eval answer-format distributions keeps SFT a fair baseline. After
  % filtering, the training set contains $1{,}771$ examples and the held-out set
  % $442$, which we split into validation ($80$) and test ($362$), stratified by
  % answer type with a fixed seed.
\begin{table}[t]
\centering
\small
\caption{Validation and test sizes for new-task and prior-capability evaluation datasets. We split each Eval dataset into validation and test subsets.}
\label{tab:dataset-splits}
\setlength{\tabcolsep}{8pt}
\begin{tabular}{lrrr}
\toprule
Dataset & \#Val & \#Test & \#Total Eval \\
\midrule
\multicolumn{4}{l}{\textbf{New-task datasets}} \\
Science & $100$ & $407$ & $507$ \\
ToolUse & -- & -- & $97$ \\
MBE & $230$ & $920$ & $1{,}150$ \\
\midrule
\multicolumn{4}{l}{\textbf{Prior-capability datasets}} \\
Code (MBPP + HumanEval) & $250$ & $414$ & $664$ \\
IFEval & $270$ & $271$ & $541$ \\
TruthfulQA-MC1 & $350$ & $467$ & $817$ \\
MMLU (57 subjects) & $503$ & $13{,}539$ & $14{,}042$ \\
HellaSwag & $500$ & $9{,}542$ & $10{,}042$ \\
WinoGrande & $300$ & $967$ & $1{,}267$ \\
MATH & $500$ & $4{,}500$ & $5{,}000$ \\
WildJailbreak & $300$ & $1{,}910$ & $2{,}210$ \\
WildGuardTest & $300$ & $1{,}425$ & $1{,}725$ \\
\bottomrule
\end{tabular}
\end{table}

\paragraph{General Language Model Capabilities Evaluation.}
We evaluate retention on nine prior-capability categories: code generation (MBPP~\cite{austin2021program} + HumanEval~\cite{chen2021evaluating}), instruction following (IFEval~\cite{zhou2023instruction}), factuality (TruthfulQA-MC1~\cite{lin2022truthfulqa}), general knowledge (MMLU~\cite{hendrycks2020measuring}), commonsense reasoning (HellaSwag~\cite{zellers2019hellaswag}), coreference reasoning (WinoGrande~\cite{sakaguchi2021winogrande}), mathematical reasoning (MATH~\cite{hendrycks2021measuring}), jailbreak robustness (WildJailbreak~\cite{jiang2024wildteaming}), and harmful-prompt refusal behavior (WildGuardTest~\cite{han2024wildguard}). We include the safety tasks because safety behavior is known to be sensitive to fine-tuning~\citep{qi2024fine}. Validation/test split sizes are given in \Cref{tab:dataset-splits}; the same fixed splits are used for every method and checkpoint.

We run HellaSwag, MMLU, WinoGrande, TruthfulQA-MC1, IFEval, MBPP, and HumanEval through the LM Evaluation Harness. For HellaSwag, MMLU, WinoGrande, TruthfulQA-MC1, and IFEval, we use fixed harness metrics: accuracy for HellaSwag, MMLU, and WinoGrande; MC1 accuracy for TruthfulQA; and instruction-level loose accuracy for IFEval, computed over the $834$ individual instructions. Because HumanEval is too small to support a stable validation split on its own, we pool it with MBPP into a single $664$-example code category, stratifying the validation/test split by source dataset. For this code category, we score generated samples with our execution-based grader. This differs from the default harness score for MBPP, which is brittle to harmless formatting differences, but gives the same score as standard \texttt{pass@1} for HumanEval. We report code as a micro-average over the pooled examples, rather than averaging the two task accuracies. MATH is graded by extracting the final boxed answer and exact-matching it to the gold answer. WildJailbreak and WildGuardTest are both evaluated by judging model responses with the WildGuard judge and scoring against the dataset labels.

We divide the above suite into two groups. The non-generative prior evaluations are MMLU, HellaSwag, Winogrande, and TruthfulQA, where the model is evaluated through constrained answer choices or short-form classification. The generative prior evaluations are IFEval, Code, Math, WildJailbreak, and WildGuardTest, which require the model to produce free-form responses that are then scored by task-specific rules or external judges.

\begin{table}[!t]
\centering
\small
\setlength{\tabcolsep}{4pt}
\renewcommand{\arraystretch}{1.15}
\begin{tabularx}{\linewidth}{
    @{}l
    >{\raggedright\arraybackslash}X
    >{\raggedright\arraybackslash}X@{}
}
\toprule
\textbf{Hyperparameter} & \textbf{SFT} & \textbf{SDFT} \\
\midrule
Base model
& Qwen2.5 7B-Instruct; Qwen2.5 14B-Instruct
& Qwen2.5 7B-Instruct; Qwen2.5 14B-Instruct \\

Learning rate (7B)
& \{5e-6, \textbf{1e-5}, \textbf{2e-5}, 3e-5, 4e-5,
   \textbf{5e-5}, \textbf{1e-4}\}
& \{5e-6, \textbf{1e-5}, \textbf{2e-5}, 3e-5, 4e-5,
   \textbf{5e-5}, \textbf{1e-4}\} \\

% Confirm the 14B ranges before submission.
Learning rate (14B)
& \{1e-5, \textbf{2e-5}, 3e-5, 4e-5,
   \textbf{5e-5}, \textbf{1e-4}\}
& \{1e-5, \textbf{2e-5}, 3e-5, 4e-5,
   \textbf{5e-5}, \textbf{1e-4}\} \\

Optimizer (7B) & AdamW & AdamW \\
Optimizer (14B) & Paged 8-bit AdamW & Paged 8-bit AdamW \\
LR scheduler & Cosine with warmup & Cosine with warmup \\
Warmup ratio & 0.1 & 0.1 \\
Epochs & 2 & 2 \\
Batch size
& \{\textbf{16}, \textbf{32}, 64\}
& \{\textbf{16}, \textbf{32}, 64\} \\
Max gradient norm & 1 & 1 \\
bfloat16 & True & True \\
Weight decay & 0 & 0 \\
Number of seeds & 3 & 3 \\
\midrule
\multicolumn{3}{l}{\emph{SDFT-only hyperparameters}} \\
EMA $\alpha$ & -- & \{0.01, \textbf{0.05}\} \\
Max generation length & -- & 2048 \\
\bottomrule
\end{tabularx}
\caption{Hyperparameter settings for SFT and SDFT. Boldface values define a smaller subsweep that can be useful under resource constraints.}
\label{tab:hyperparameters}
\end{table}

Unless otherwise stated, aggregate retention is the unweighted mean over all of the above prior-capability categories.

\subsection{Hyperparameter Choices}\label{subsec:hp-choices}

We sweep the values in \Cref{tab:hyperparameters}. Hyperparameter selection uses validation scores averaged across three seeds. Whenever we examine the effect of a subset of hyperparameters, we select the remaining hyperparameters according to the highest joint utility \eqref{eq:utility-LR}.

\paragraph{Optimizer.}
All Qwen2.5-7B experiments use AdamW. For Qwen2.5-14B, we use paged 8-bit AdamW for memory efficiency.

\paragraph{Compute.}
Each run uses one NVIDIA B200 GPU with \texttt{bfloat16} precision and gradient checkpointing. We use a per-device batch size of $1$ and gradient accumulation to obtain the reported batch sizes. We estimate total compute for the study at approximately $3{,}000$ GPU-hours.

% ---------------------------------------------------------------- cost comparison
\paragraph{Compute cost of SDFT.}
SDFT is more expensive than SFT because it generates on-policy student rollouts. We estimate that SDFT costs roughly $1.3\times$ as much as SFT end-to-end and $1.9\times$ for training, on average across Qwen2.5-7B and Qwen2.5-14B. The smaller end-to-end difference reflects the evaluation overhead shared by both methods. SDFT may also require additional runs to tune the teacher EMA parameter $\alpha$, which is important for achieving its best performance.

% ---------------------------------------------------------------- optional table

\section{Deferred Details\texorpdfstring{ from \Cref{sec:main-results}}{} (Qwen 2.5-7B)}
\label{app:qwen7b-main-results}
\suppressfloats[t]

This appendix provides additional results for the Qwen2.5-7B findings in \Cref{sec:main-results}. To complement the aggregate comparison in \Cref{fig:qwen7b-final-comparison}, \Cref{tab:qwen7b_best} compares scores on individual benchmarks with the base model for the configurations selected by joint utility. For example, after fine-tuning on MBE, MATH drops by about 19 pp with SFT and 2 pp with SDFT, while IFEval drops by about 23 pp with either method.

\begin{figure}[H]
    \centering
    \includegraphics[width=\textwidth]{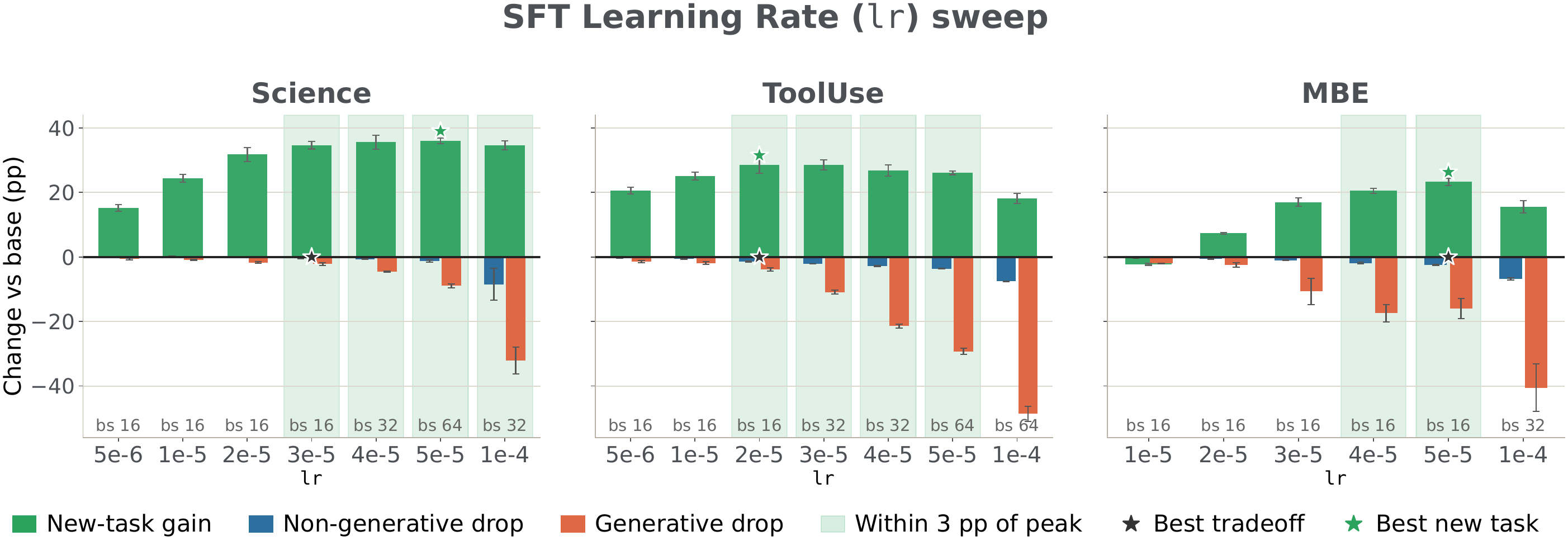}
    \includegraphics[width=\textwidth]{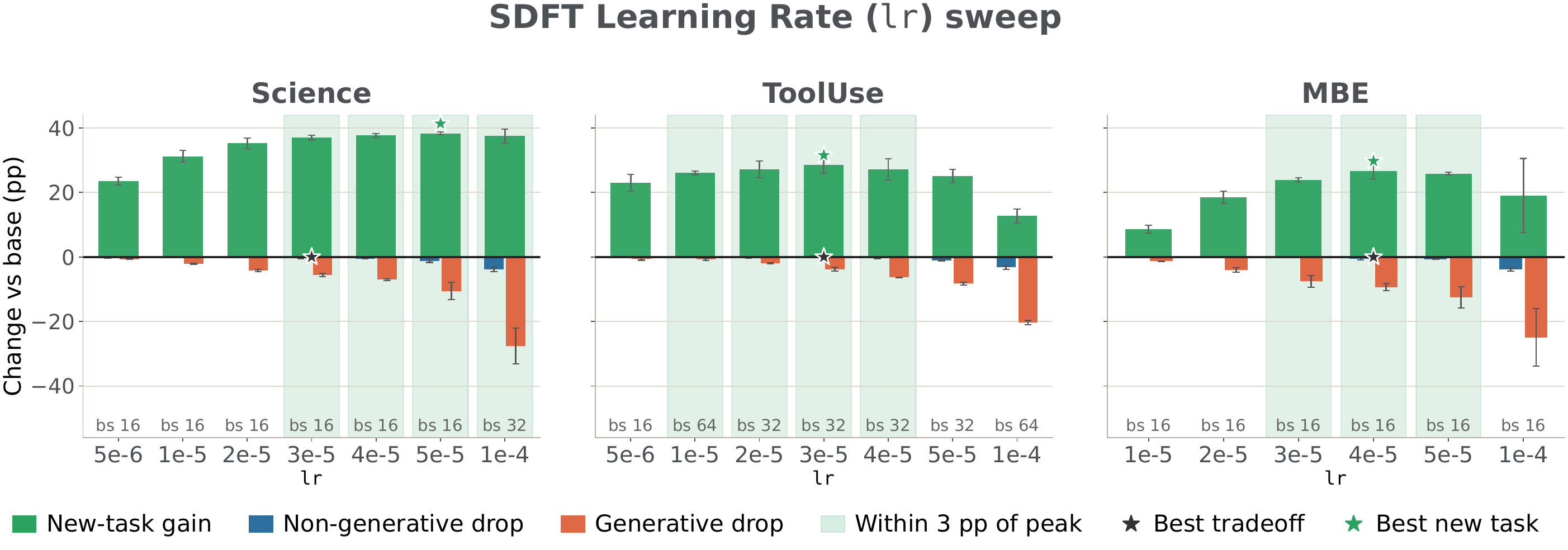}
    \caption{\textbf{Generative evaluations show the largest drops at high learning rates.}
    Learning-rate sweeps for Qwen2.5-7B with SFT (top) and SDFT (bottom), using the same configurations as \Cref{fig:qwen7b-lr-sweep-gain-drop}.
    Bars show changes from the base model on the new task (green) and unweighted mean changes within the generative (orange) and non-generative (blue) benchmark groups.}
    \label{fig:qwen7b-lr-sweep-drop-split}
\end{figure}

\begin{table}[!htbp]
\centering
\caption{Per-capability performance for Qwen2.5-7B configurations
selected by joint utility \eqref{eq:utility-LR}. The selected SDFT
configurations use EMA parameter $\alpha=0.05$ for Science and MBE and
$\alpha=0.01$ for ToolUse. The final column reports the unweighted
average accuracy across the nine prior-capability categories. Subscripts
indicate standard deviations.}
\label{tab:qwen7b_best}
\resizebox{\linewidth}{!}{%
\begin{tabular}{llcccccccccc|c}
\toprule
Task & Method & New task & MMLU & HellaSw. & Wino. & TQA1 & IFEval & Code & MATH & WJB & WGT & \textsc{Avg} \\
\midrule
Science
& Base & 33.93 & 71.76 & 61.99 & 70.96 & 47.86 & 73.26 & 60.09 & 72.26 & 21.54 & 77.22 & $\mathbf{61.88}$ \\
& SFT, 3e-5/bs16 & $68.57_{\pm 1.16}$ & 71.48 & 61.36 & 72.67 & 45.41 & 64.23 & 65.91 & 70.19 & 17.98 & 75.13 & $\mathbf{60.48}_{\pm 0.27}$ \\
& SDFT, 3e-5/bs16 & $70.94_{\pm 0.75}$ & 71.38 & 60.95 & 72.74 & 46.72 & 57.67 & 62.15 & 70.19 & 13.44 & 72.95 & $\mathbf{58.69}_{\pm 0.24}$ \\
\midrule
ToolUse
& Base & 41.24 & 71.76 & 61.99 & 70.96 & 47.86 & 73.26 & 60.09 & 72.26 & 21.54 & 77.22 & $\mathbf{61.88}$ \\
& SFT, 2e-5/bs16 & $69.76_{\pm 2.59}$ & 71.51 & 60.49 & 72.93 & 42.06 & 65.83 & 58.48 & 69.41 & 17.63 & 73.76 & $\mathbf{59.12}_{\pm 0.39}$ \\
& SDFT, 3e-5/bs32 & $69.76_{\pm 2.59}$ & 71.50 & 61.18 & 73.01 & 46.96 & 60.59 & 63.75 & 70.96 & 15.78 & 74.34 & $\mathbf{59.78}_{\pm 0.40}$ \\
\midrule
MBE
& Base & 49.74 & 71.76 & 61.99 & 70.96 & 47.86 & 73.26 & 60.09 & 72.26 & 21.54 & 77.22 & $\mathbf{61.88}$ \\
& SFT, 5e-5/bs16 & $73.04_{\pm 1.15}$ & 71.55 & 60.88 & 71.20 & 39.29 & 50.40 & 45.38 & 52.89 & 10.68 & 65.22 & $\mathbf{51.94}_{\pm 1.68}$ \\
& SDFT, 4e-5/bs16 & $76.46_{\pm 2.52}$ & 71.68 & 60.98 & 72.58 & 45.41 & 49.96 & 55.07 & 70.01 & 11.76 & 70.92 & $\mathbf{56.49}_{\pm 0.59}$ \\
\bottomrule
\end{tabular}}
\end{table}
\paragraph{Most of the drop is generative.}
Recall from \Cref{sec:dataset-for-us} that our generative evaluations score
free-form responses for instruction following (IFEval), code generation
(MBPP and HumanEval), mathematical reasoning (MATH), jailbreak robustness
(WildJailbreak), and harmful-prompt refusal (WildGuardTest).
\Cref{fig:qwen7b-lr-sweep-drop-split} revisits the learning-rate sweep in
\Cref{fig:qwen7b-lr-sweep-gain-drop}, reporting retention separately for
these benchmarks and for the non-generative benchmarks. At the highest
learning rate, the generative group drops by about 20--49 pp across the
three tasks and both methods, compared with 3--8 pp for the non-generative
group.

\begin{figure}[H]
    \centering
    \includegraphics[width=.85\linewidth]{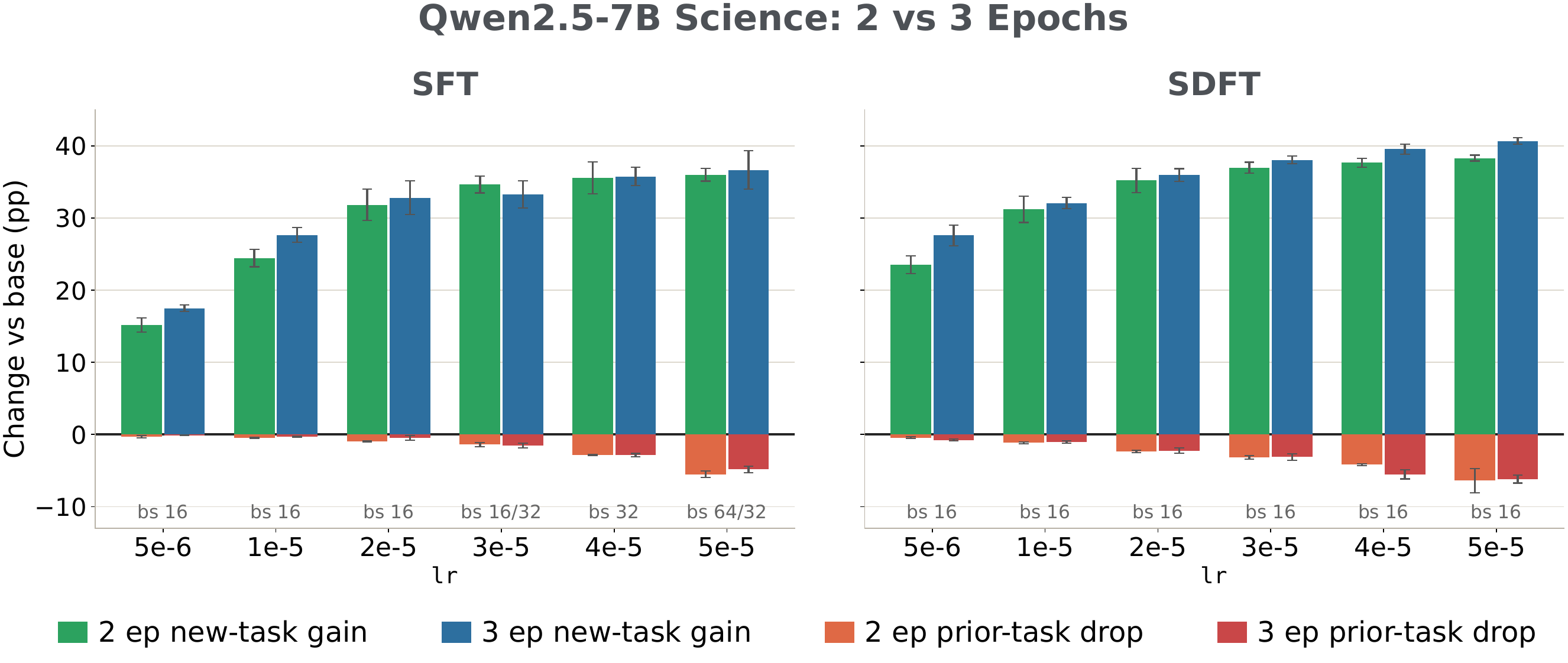}
    \caption{A third epoch at low learning rates yields modest Science gains with little change in retention. In this sweep, matching the best two-epoch gains requires a higher learning rate; training longer at a lower rate does not close the gap.}
    \label{fig:qwen7b-science-2-vs-3-ep}
\end{figure}

\paragraph{Effect of the number of epochs.}
Following \Cref{subsec:additional-obs}, \Cref{fig:qwen7b-science-2-vs-3-ep} tests whether an extra epoch at a lower learning rate can match the gains from a higher rate without the additional forgetting.

\begin{figure}[H]
    \centering
    \includegraphics[width=\linewidth]{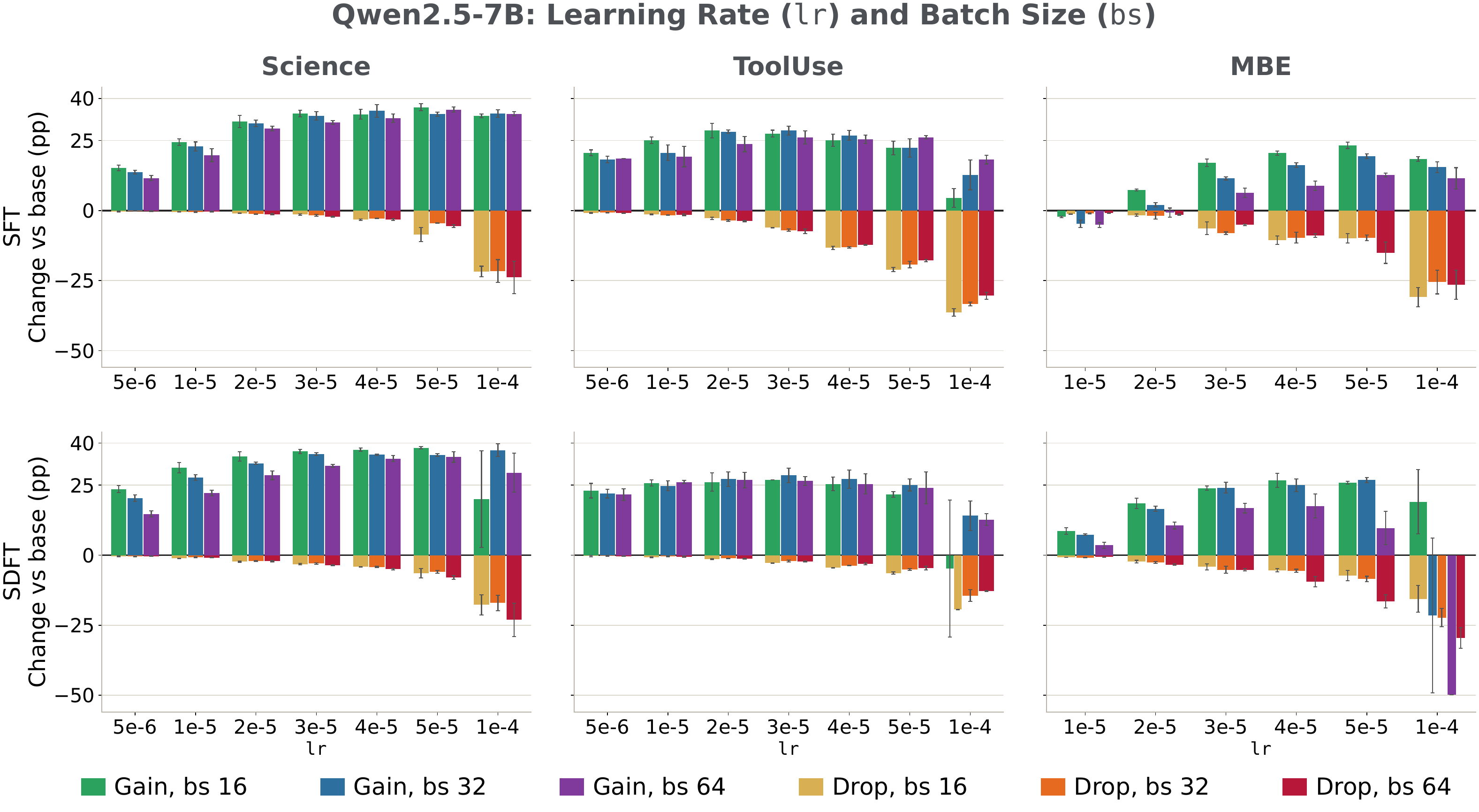}
    \caption{The gains and drop for different batch sizes at different learning rates.}
    \label{fig:lr-bs-interaction}
\end{figure}

\paragraph{Effect of batch size.} \Cref{fig:lr-bs-interaction} shows all three batch sizes for the learning-rate sweep in \Cref{fig:qwen7b-lr-sweep-gain-drop}. Once forgetting becomes catastrophic, the effect of batch size varies across tasks and methods, but we highlight one clean observation at lower learning rates. Below the onset of catastrophic forgetting up to a critical learning rate, batch size 16 generally nearly matches or improves on the new-task gains of larger batches without worsening retention. At a fixed epoch count, smaller batches give more optimizer updates, which may explain the stronger gains.

\clearpage
\section{Results at Larger Scale on Qwen2.5-14B}\label{app:14b}

\begin{figure}[!ht]
    \centering
    \includegraphics[width=\linewidth]
        {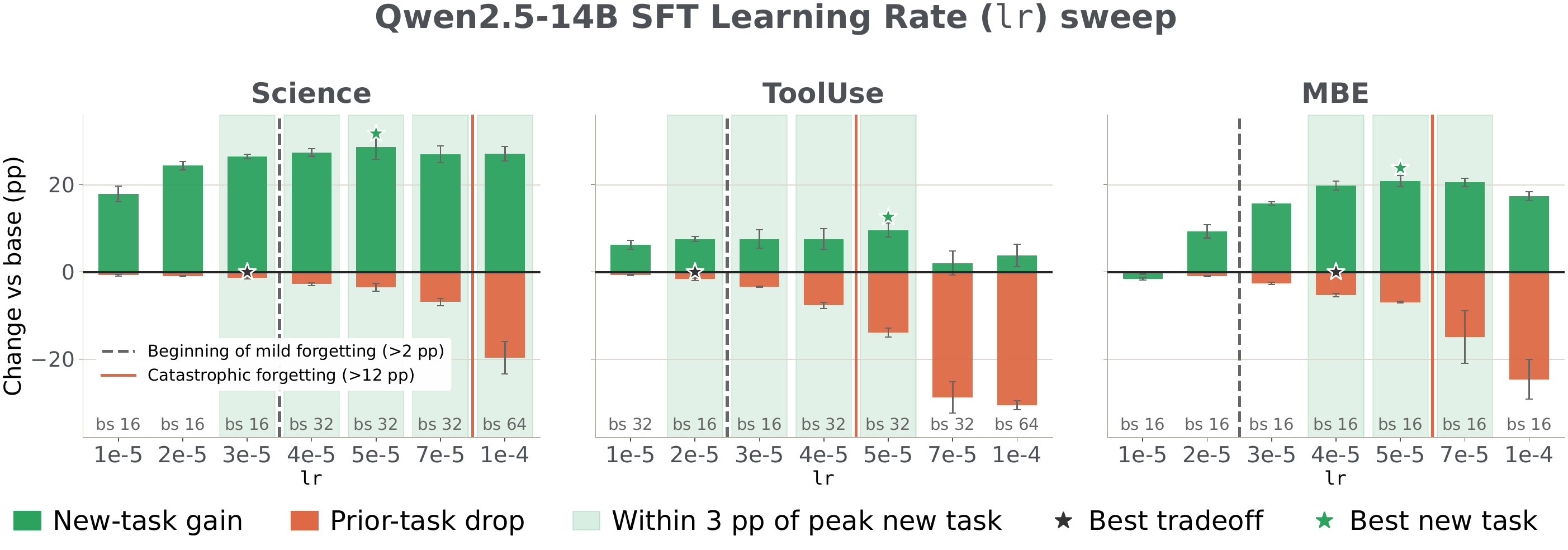}

    \vspace{1mm}
    \includegraphics[width=\linewidth]
        {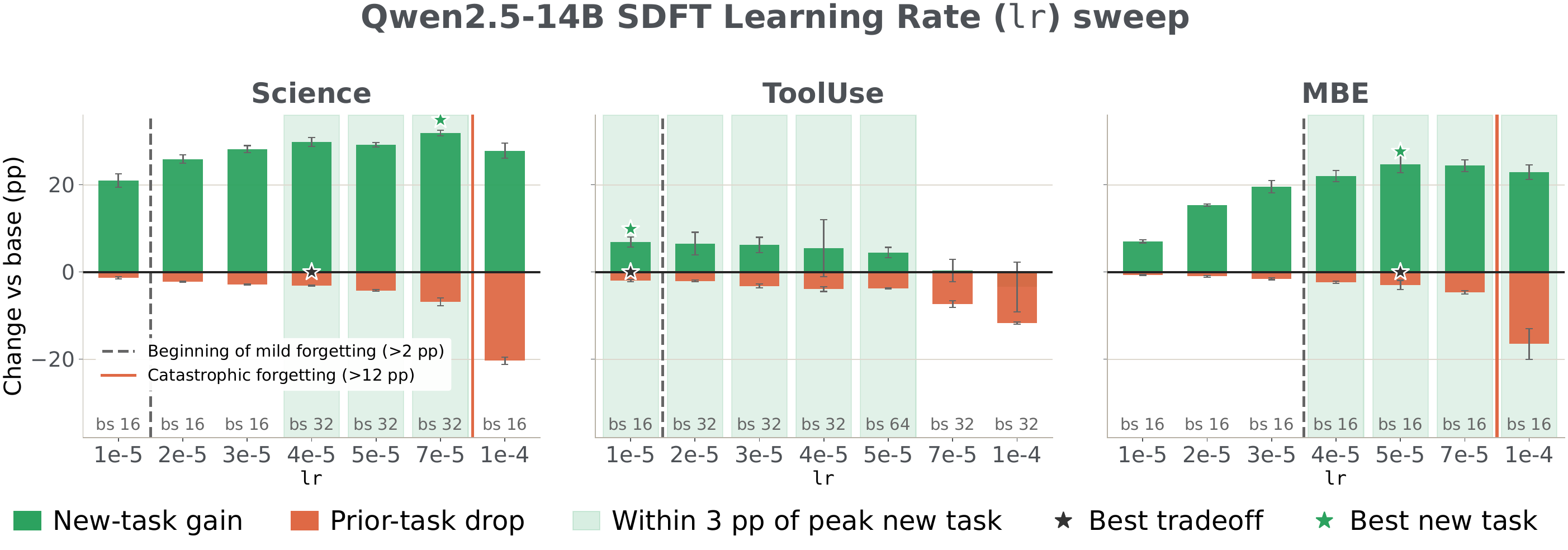}
\vspace{-20pt}
    \caption{Learning-rate sweeps for Qwen2.5-14B with SFT (top)
    and SDFT (bottom). At each learning rate, we select the batch size
    that maximizes joint utility \eqref{eq:utility-LR}, with 2 epochs of training. Bars show new-task gains and average
    prior-task drops relative to the base model.}
    \label{fig:qwen14b-lr-sweep-gain-drop}
\end{figure}

\Cref{fig:qwen14b-lr-sweep-gain-drop} shows that the main effects persist
at 14B (recall the discussion in \Cref{subsec:lr-learning-forgetting}),
with learning rate remaining a primary driver of the learning--retention
tradeoff for both SFT and SDFT. While SDFT again exhibits greater robustness,
with less severe drops in some settings, it too can yield checkpoints
with similar gains but forgetting ranging from catastrophic to negligible.

\begin{figure}[!h]
    \centering
    \includegraphics[width=\linewidth]
        {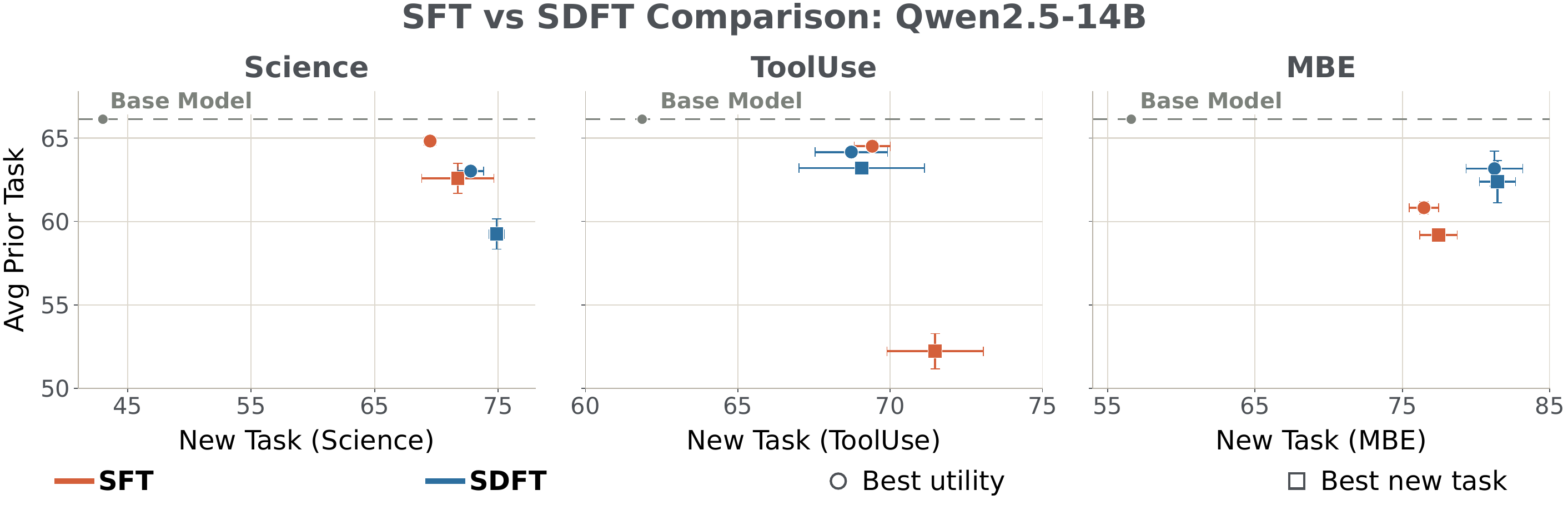}
        \vspace{-20pt}
    \caption{SFT vs SDFT comparison for
    Qwen2.5-14B.}
    \vspace{-20pt}
    \label{fig:qwen14b-final-comparison}
\end{figure}

We repeat the comparison on Qwen2.5-14B using the hyperparameter sweep in
\Cref{tab:hyperparameters}.
\Cref{fig:qwen14b-final-comparison} and \Cref{tab:qwen14b_best} show that
appropriately tuned SFT remains competitive, particularly on Science
and ToolUse. Under joint-utility selection \eqref{eq:utility-LR}, SFT achieves somewhat
smaller Science gains but also forgets less, while slightly exceeding
SDFT's mean ToolUse performance with better retention. Selecting only
for new-task gain can substantially worsen retention for modest
additional gains, most visibly for SFT on ToolUse. When greater
new-task learning involves trading off prior capabilities, as in MBE,
SDFT again offers an advantage through less severe retention losses.
\begin{table}[!ht]
\centering
\caption{Per-capability performance for Qwen2.5-14B configurations
selected by joint utility \eqref{eq:utility-LR}. All selected SDFT
configurations use EMA parameter $\alpha=0.05$. The final column reports the
unweighted average accuracy across the nine prior-capability categories. Subscripts
indicate standard deviations.}
\label{tab:qwen14b_best}
\resizebox{\linewidth}{!}{%
\begin{tabular}{llcccccccccc|c}
\toprule
Task & Method & New task & MMLU & HellaSw. & Wino. & TQA1
& IFEval & Code & MATH & WJB & WGT & Avg. accuracy \\
\midrule
Science
& Base & 43.00
& 78.87 & 65.62 & 75.85 & 51.77 & 70.38 & 63.86 & 76.16 & 32.81 & 79.83
& $\mathbf{66.13}$ \\
& SFT, 3e-5/bs16 & $69.49_{\pm 0.50}$
& 77.80 & 65.07 & 76.35 & 48.31 & 66.03 & 72.04 & 75.03 & 24.90 & 77.86
& $\mathbf{64.82}_{\pm 0.25}$ \\
& SDFT, 4e-5/bs32 & $72.78_{\pm 1.04}$
& 77.93 & 64.78 & 77.74 & 49.81 & 56.75 & 71.19 & 74.45 & 19.01 & 75.52
& $\mathbf{63.02}_{\pm 0.14}$ \\
\midrule
ToolUse
& Base & 61.86
& 78.87 & 65.62 & 75.85 & 51.77 & 70.38 & 63.86 & 76.16 & 32.81 & 79.83
& $\mathbf{66.13}$ \\
& SFT, 2e-5/bs16 & $69.41_{\pm 0.59}$
& 78.25 & 64.44 & 76.95 & 48.14 & 66.67 & 67.37 & 73.97 & 26.59 & 78.18
& $\mathbf{64.51}_{\pm 0.34}$ \\
& SDFT, 1e-5/bs16 & $68.73_{\pm 1.19}$
& 78.64 & 65.31 & 75.80 & 49.41 & 66.71 & 65.66 & 75.11 & 23.21 & 77.60
& $\mathbf{64.16}_{\pm 0.28}$ \\
\midrule
MBE
& Base & 56.61
& 78.87 & 65.62 & 75.85 & 51.77 & 70.38 & 63.86 & 76.16 & 32.81 & 79.83
& $\mathbf{66.13}$ \\
& SFT, 4e-5/bs16 & $76.46_{\pm 1.00}$
& 78.65 & 65.94 & 75.64 & 42.96 & 68.15 & 70.89 & 56.57 & 15.16 & 73.45
& $\mathbf{60.82}_{\pm 0.38}$ \\
& SDFT, 5e-5/bs16 & $81.25_{\pm 1.93}$
& 77.84 & 64.86 & 77.85 & 48.92 & 63.67 & 71.64 & 72.21 & 16.70 & 74.82
& $\mathbf{63.17}_{\pm 1.05}$ \\
\bottomrule
\end{tabular}}
\end{table}

\section{Additional Correlations and Model-Family Results}\label{sec:remaining-details}

\begin{figure}[!ht]
\centering
\includegraphics[width=0.36\linewidth]{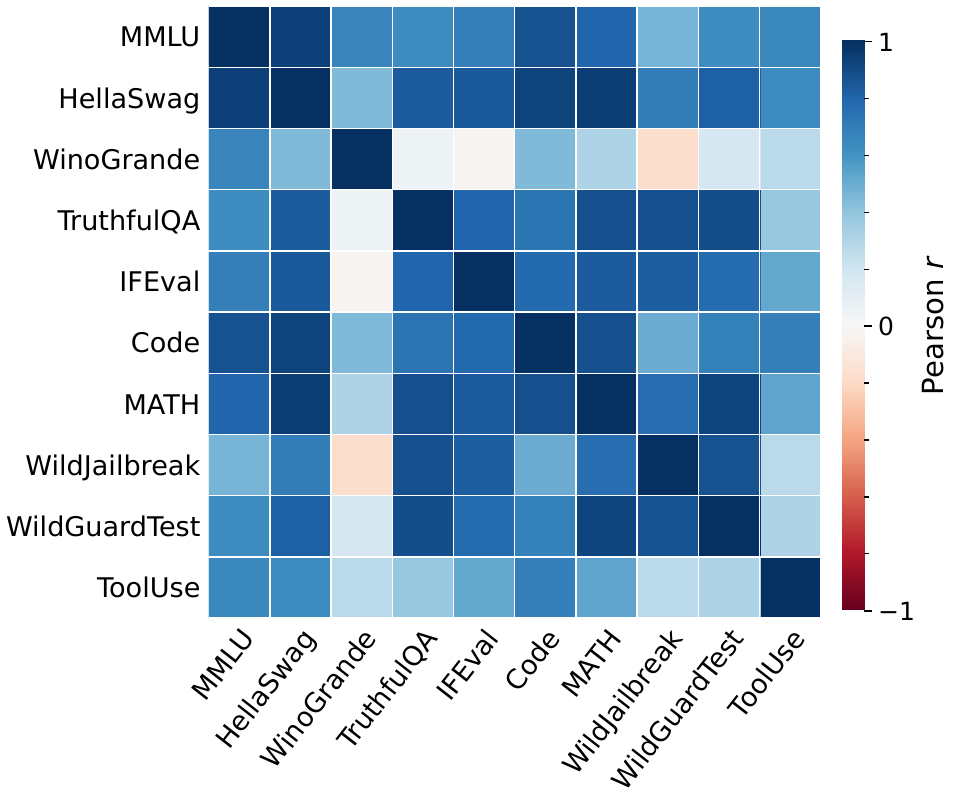}\hspace{0.02\linewidth}%
\includegraphics[width=0.36\linewidth]{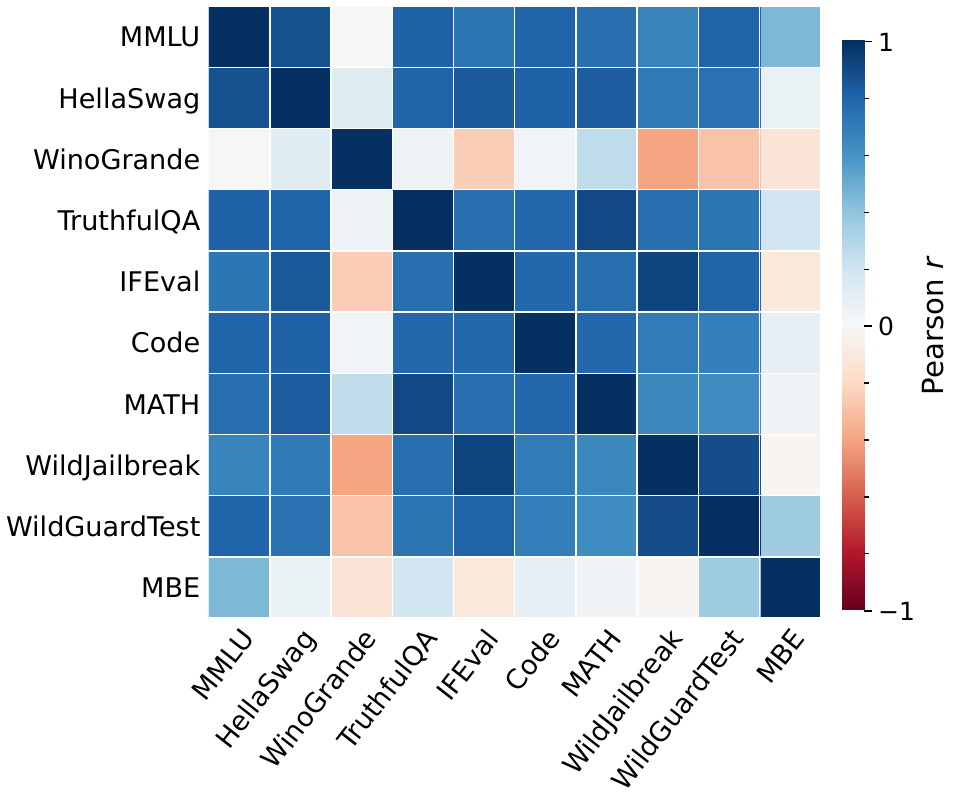}
\captionof{figure}{Capability-change correlations for Qwen2.5-7B:
ToolUse (left) and MBE (right). Pearson correlations are computed across
 two-epoch SFT and SDFT configurations.}
\label{fig:7b-correlations}
\end{figure}

\begin{figure}[H]
\centering
\includegraphics[width=0.328\linewidth]{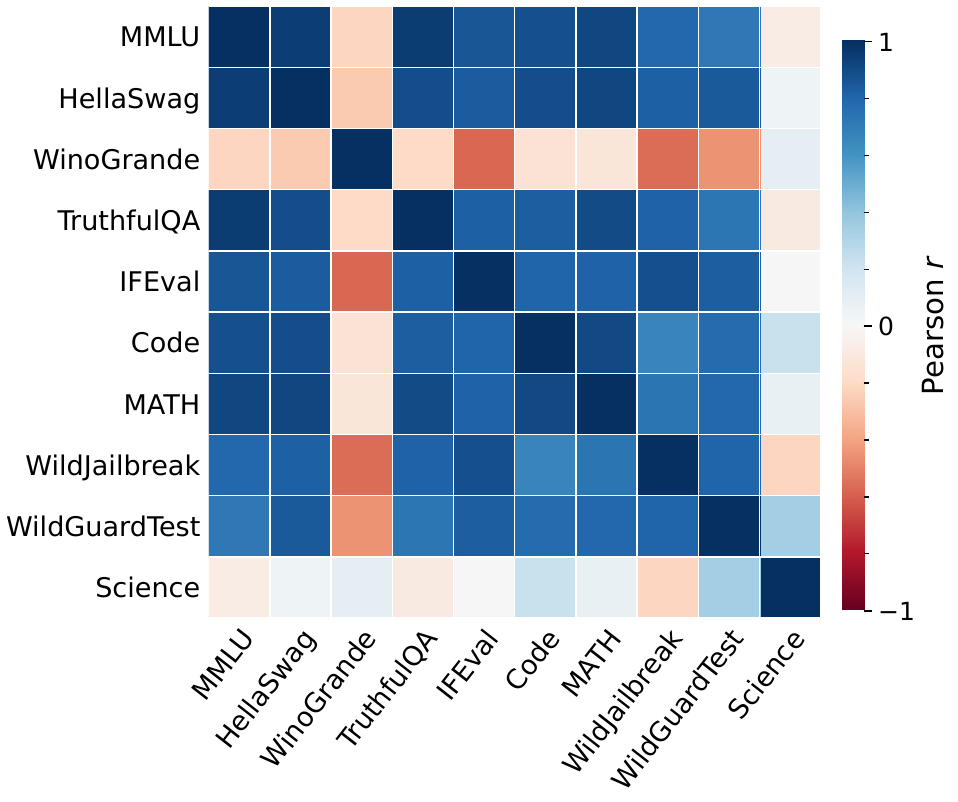}\hfill%
\includegraphics[width=0.328\linewidth]{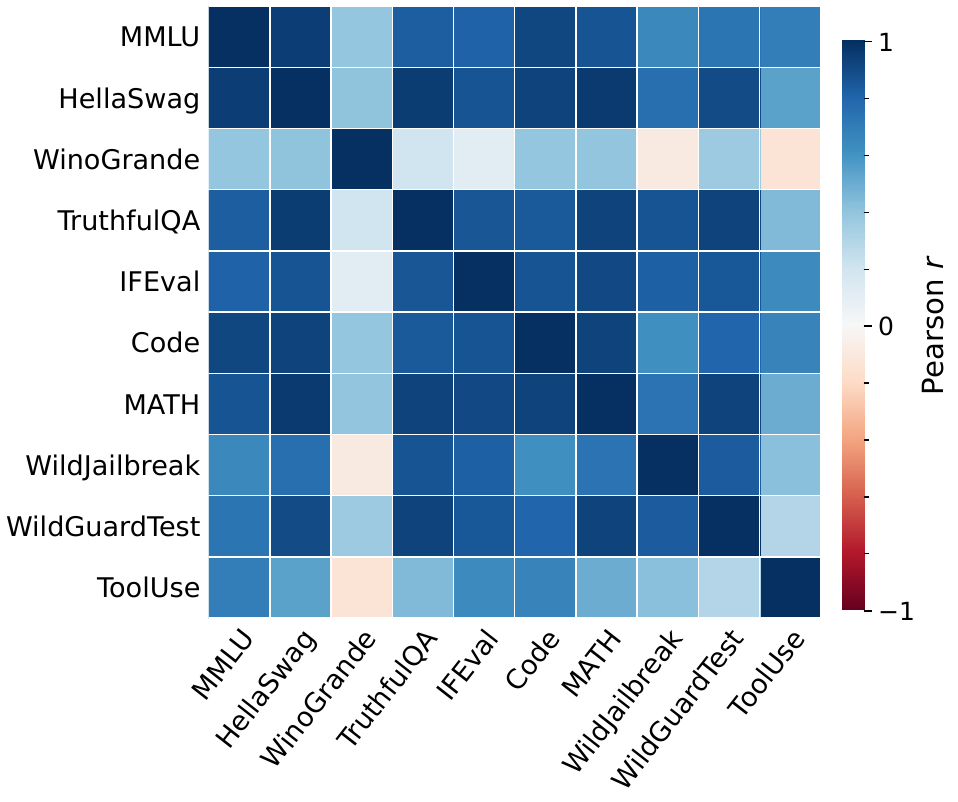}\hfill%
\includegraphics[width=0.328\linewidth]{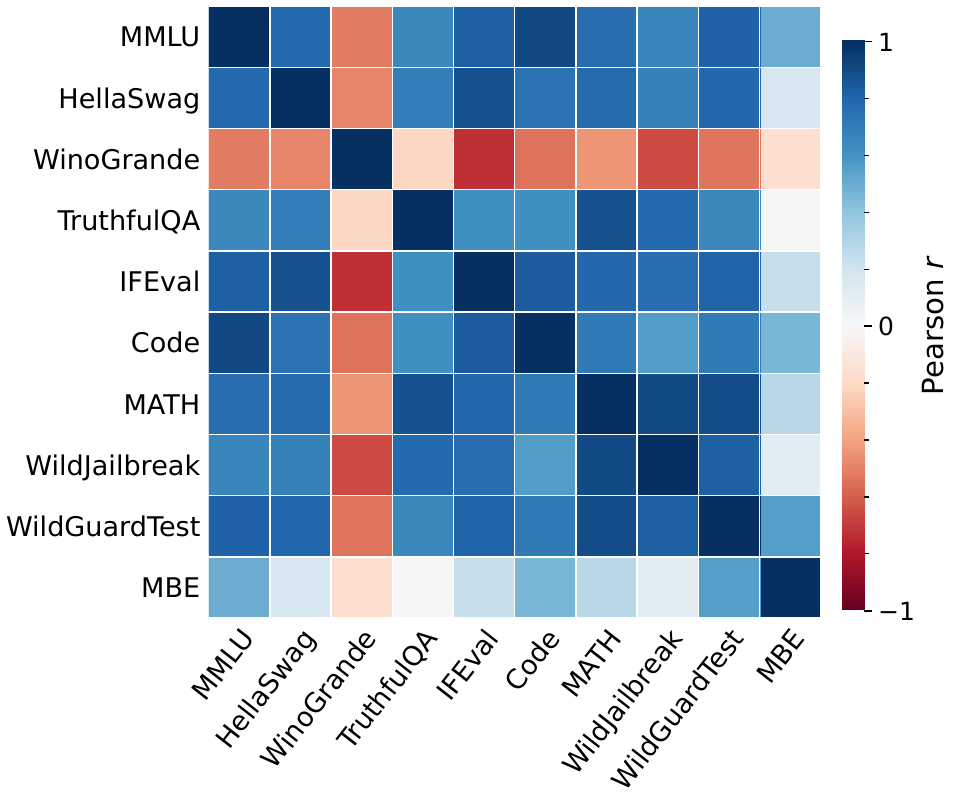}
\caption{Capability-change correlations for Qwen2.5-14B:
Science (left), ToolUse (middle), and MBE (right), computed as above
across $34$ configurations per task.}
\label{fig:14b-correlations}
\end{figure}

We provide further evidence that drops in prior capabilities are highly
correlated as we fine-tune on a new task. \Cref{fig:7b-correlations}
complements the Science results in \Cref{fig:science-capability-correlations}
with ToolUse and MBE at 7B; \Cref{fig:14b-correlations} shows the
corresponding patterns for all three tasks at 14B.


\begin{figure}[H]
\centering
\includegraphics[width=0.85\linewidth]{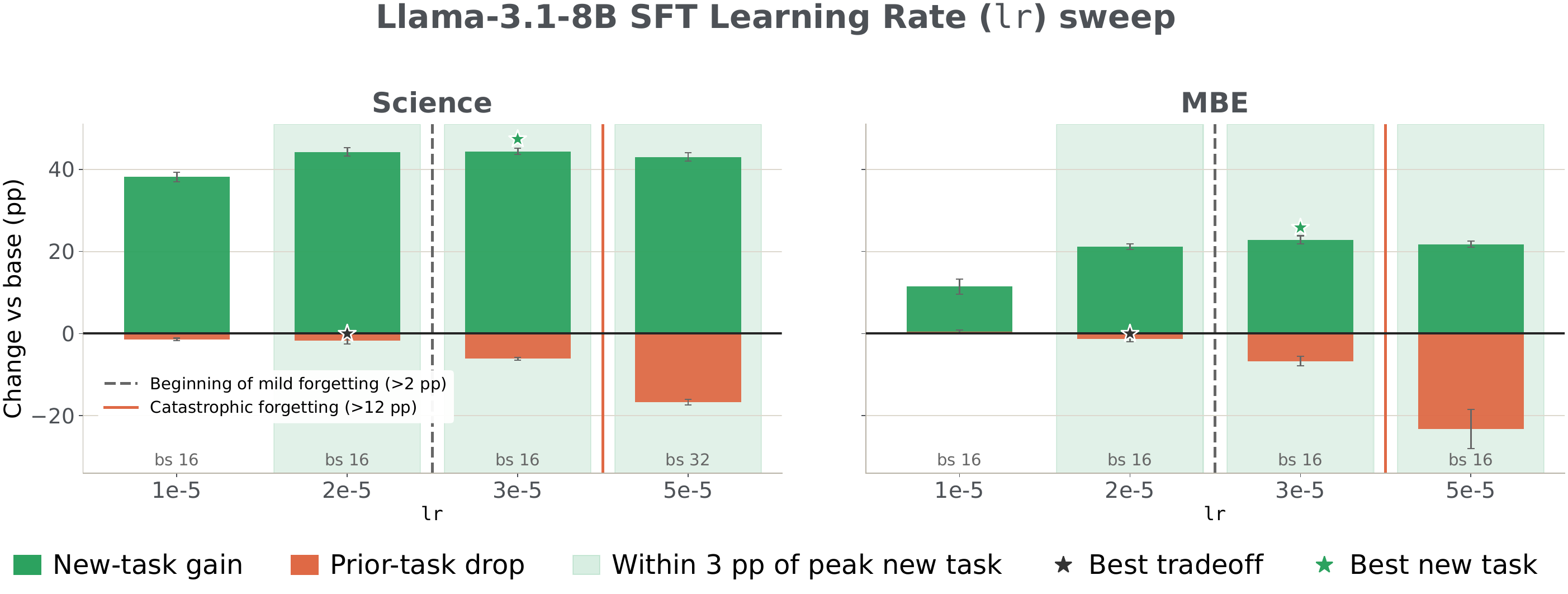}
\caption{Learning-rate sweep for SFT on Llama-3.1-8B.
At each learning rate, we select the batch size maximizing joint utility
\eqref{eq:utility-LR}, with two epochs of training. Bars show new-task
gains and average prior-task drops; error bars show standard deviations.}
\label{fig:llama8b-lr-sweep-gain-drop}
\end{figure}

We also run SFT on Llama-3.1-8B and observe the same main effects along
the learning-rate axis: larger learning rates can improve new-task gains
but lead to substantial prior-capability drops
(\Cref{fig:llama8b-lr-sweep-gain-drop}).

%% file: main.bbl
\begin{thebibliography}{29}
\providecommand{\natexlab}[1]{#1}
\providecommand{\url}[1]{\texttt{#1}}
\expandafter\ifx\csname urlstyle\endcsname\relax
  \providecommand{\doi}[1]{doi: #1}\else
  \providecommand{\doi}{doi: \begingroup \urlstyle{rm}\Url}\fi

\bibitem[Agarwal et~al.(2024)Agarwal, Vieillard, Zhou, Stanczyk, Ramos~Garea, Geist, and Bachem]{agarwal2024policy}
Rishabh Agarwal, Nino Vieillard, Yongchao Zhou, Piotr Stanczyk, Sabela Ramos~Garea, Matthieu Geist, and Olivier Bachem.
\newblock On-policy distillation of language models: Learning from self-generated mistakes.
\newblock In \emph{International Conference on Learning Representations}, volume 2024, pages 21246--21263, 2024.

\bibitem[Austin et~al.(2021)Austin, Odena, Nye, Bosma, Michalewski, Dohan, Jiang, Cai, Terry, Le, et~al.]{austin2021program}
Jacob Austin, Augustus Odena, Maxwell Nye, Maarten Bosma, Henryk Michalewski, David Dohan, Ellen Jiang, Carrie Cai, Michael Terry, Quoc Le, et~al.
\newblock Program synthesis with large language models.
\newblock \emph{arXiv preprint arXiv:2108.07732}, 2021.

\bibitem[Biderman et~al.(2024)Biderman, Portes, Ortiz, Paul, Greengard, Jennings, King, Havens, Chiley, Frankle, et~al.]{biderman2024lora}
Dan Biderman, Jacob Portes, Jose Javier~Gonzalez Ortiz, Mansheej Paul, Philip Greengard, Connor Jennings, Daniel King, Sam Havens, Vitaliy Chiley, Jonathan Frankle, et~al.
\newblock Lora learns less and forgets less.
\newblock \emph{arXiv preprint arXiv:2405.09673}, 2024.

\bibitem[Chen et~al.(2026)Chen, Razin, Narasimhan, and Chen]{chen2026retaining}
Howard Chen, Noam Razin, Karthik Narasimhan, and Danqi Chen.
\newblock Retaining by doing: The role of on-policy data in mitigating forgetting.
\newblock In \emph{International Conference on Machine Learning}, 2026.

\bibitem[Chen et~al.(2021)Chen, Tworek, Jun, Yuan, de~Oliveira~Pinto, Kaplan, Edwards, Burda, Joseph, Brockman, Ray, Puri, Krueger, Petrov, Khlaaf, Sastry, Mishkin, Chan, Gray, Ryder, Pavlov, Power, Kaiser, Bavarian, Winter, Tillet, Such, Cummings, Plappert, Chantzis, Barnes, Herbert-Voss, Guss, Nichol, Paino, Tezak, Tang, Babuschkin, Balaji, Jain, Saunders, Hsu, Crowder, Srinivasan, Ho, Gray, Ryder, Amodei, Sutskever, and Zaremba]{chen2021evaluating}
Mark Chen, Jerry Tworek, Heewoo Jun, Qiming Yuan, Henrique~Pond{\'e} de~Oliveira~Pinto, Jared Kaplan, Harri Edwards, Yuri Burda, Nicholas Joseph, Greg Brockman, Alex Ray, Sandhini Puri, Gretchen Krueger, Mikhail Petrov, Heidy Khlaaf, Girish Sastry, Pamela Mishkin, Brooke Chan, Scott Gray, Nick Ryder, Michael Pavlov, Alethea Power, Lukasz Kaiser, Mohammad Bavarian, Clemens Winter, Phil Tillet, Felipe Such, Dave Cummings, Matthias Plappert, Fotios Chantzis, Elizabeth Barnes, Ariel Herbert-Voss, William~H. Guss, Alex Nichol, Igor Paino, Nikolas Tezak, Jie Tang, Igor Babuschkin, Suchir Balaji, Szymon Jain, William Saunders, Jesse Hsu, Ryan Crowder, Anjali Srinivasan, Andrew Ho, Tom Gray, Nicholas Ryder, Dario Amodei, Ilya Sutskever, and Wojciech Zaremba.
\newblock Evaluating large language models trained on code.
\newblock \emph{arXiv preprint arXiv:2107.03374}, 2021.

\bibitem[Ding and Wang(2025)]{ding2025improved}
Fei Ding and Baiqiao Wang.
\newblock Improved supervised fine-tuning for large language models to mitigate catastrophic forgetting.
\newblock \emph{arXiv preprint arXiv:2506.09428}, 2025.

\bibitem[Feng et~al.(2024)Feng, Shen, Wang, Zhuang, Tang, Zhang, and Ding]{feng2024sciknoweval}
Kehua Feng, Xinyi Shen, Weijie Wang, Xiang Zhuang, Yuqi Tang, Qiang Zhang, and Keyan Ding.
\newblock Sciknoweval: Evaluating multi-level scientific knowledge of large language models.
\newblock \emph{arXiv preprint arXiv:2406.09098}, 2024.

\bibitem[Fernandes et~al.(2025)Fernandes, Biedenkapp, Hutter, and Awad]{fernandes2025llamawalksbarefficient}
Rean Fernandes, Andr{\'e} Biedenkapp, Frank Hutter, and Noor Awad.
\newblock A llama walks into the'bar': Efficient supervised fine-tuning for legal reasoning in the multi-state bar exam.
\newblock \emph{arXiv preprint arXiv:2504.04945}, 2025.

\bibitem[Han et~al.(2024)Han, Rao, Ettinger, Jiang, Lin, Lambert, Choi, and Dziri]{han2024wildguard}
Seungju Han, Kavel Rao, Allyson Ettinger, Liwei Jiang, Bill~Yuchen Lin, Nathan Lambert, Yejin Choi, and Nouha Dziri.
\newblock Wildguard: Open one-stop moderation tools for safety risks, jailbreaks, and refusals of llms.
\newblock \emph{Advances in neural information processing systems}, 37:\penalty0 8093--8131, 2024.

\bibitem[Hendrycks et~al.(2020)Hendrycks, Burns, Basart, Zou, Mazeika, Song, and Steinhardt]{hendrycks2020measuring}
Dan Hendrycks, Collin Burns, Steven Basart, Andy Zou, Mantas Mazeika, Dawn Song, and Jacob Steinhardt.
\newblock Measuring massive multitask language understanding.
\newblock \emph{arXiv preprint arXiv:2009.03300}, 2020.

\bibitem[Hendrycks et~al.(2021)Hendrycks, Burns, Kadavath, Arora, Basart, Tang, Song, and Steinhardt]{hendrycks2021measuring}
Dan Hendrycks, Collin Burns, Saurav Kadavath, Akul Arora, Steven Basart, Eric Tang, Dawn Song, and Jacob Steinhardt.
\newblock Measuring mathematical problem solving with the math dataset.
\newblock \emph{arXiv preprint arXiv:2103.03874}, 2021.

\bibitem[Huang et~al.(2024)Huang, Cui, Wang, Yang, Liao, Song, Yao, and Su]{huang2024mitigating}
Jianheng Huang, Leyang Cui, Ante Wang, Chengyi Yang, Xinting Liao, Linfeng Song, Junfeng Yao, and Jinsong Su.
\newblock Mitigating catastrophic forgetting in large language models with self-synthesized rehearsal.
\newblock In \emph{Proceedings of the 62nd Annual Meeting of the Association for Computational Linguistics (Volume 1: Long Papers)}, pages 1416--1428, 2024.

\bibitem[Jiang et~al.(2024)Jiang, Rao, Han, Ettinger, Brahman, Kumar, Mireshghallah, Lu, Sap, Choi, et~al.]{jiang2024wildteaming}
Liwei Jiang, Kavel Rao, Seungju Han, Allyson Ettinger, Faeze Brahman, Sachin Kumar, Niloofar Mireshghallah, Ximing Lu, Maarten Sap, Yejin Choi, et~al.
\newblock Wildteaming at scale: From in-the-wild jailbreaks to (adversarially) safer language models.
\newblock \emph{Advances in Neural Information Processing Systems}, 37:\penalty0 47094--47165, 2024.

\bibitem[Li et~al.(2024)Li, Ding, Fang, and Tao]{li2024revisiting}
Hongyu Li, Liang Ding, Meng Fang, and Dacheng Tao.
\newblock Revisiting catastrophic forgetting in large language model tuning.
\newblock In \emph{Findings of the Association for Computational Linguistics: EMNLP 2024}, pages 4297--4308. Association for Computational Linguistics, 2024.

\bibitem[Lin et~al.(2026)Lin, Wang, Qian, Wang, Srinivasan, Zeng, Jiao, Zhou, Gesi, Wang, et~al.]{lin2026sft}
Jiacheng Lin, Zhongruo Wang, Kun Qian, Tian Wang, Arvind Srinivasan, Hansi Zeng, Ruochen Jiao, Xie Zhou, Jiri Gesi, Dakuo Wang, et~al.
\newblock Sft doesn’t always hurt general capabilities: Revisiting domain-specific fine-tuning in llms.
\newblock In \emph{International Conference on Learning Representations}, volume 2026, pages 46954--46992, 2026.

\bibitem[Lin et~al.(2022)Lin, Hilton, and Evans]{lin2022truthfulqa}
Stephanie Lin, Jacob Hilton, and Owain Evans.
\newblock Truthfulqa: Measuring how models mimic human falsehoods.
\newblock In \emph{Proceedings of the 60th annual meeting of the association for computational linguistics (volume 1: long papers)}, pages 3214--3252, 2022.

\bibitem[Luo et~al.(2023)Luo, Yang, Meng, Li, Zhou, and Zhang]{luo2023empirical}
Yun Luo, Zhen Yang, Fandong Meng, Yafu Li, Jie Zhou, and Yue Zhang.
\newblock An empirical study of catastrophic forgetting in large language models during continual fine-tuning.
\newblock \emph{arXiv preprint arXiv:2308.08747}, 2023.

\bibitem[Marek et~al.(2026)Marek, Cho, Qiu, Chunara, Izmailov, and Wilson]{marek2026forgetting}
Martin Marek, Dongkyu Cho, Shikai Qiu, Rumi Chunara, Pavel Izmailov, and Andrew~Gordon Wilson.
\newblock Forgetting in language models: Capacity, optimization, and self-generated replay.
\newblock \emph{arXiv preprint arXiv:2605.26097}, 2026.

\bibitem[Qi et~al.(2024)Qi, Zeng, Xie, Chen, Jia, Mittal, and Henderson]{qi2024fine}
Xiangyu Qi, Yi~Zeng, Tinghao Xie, Pin-Yu Chen, Ruoxi Jia, Prateek Mittal, and Peter Henderson.
\newblock Fine-tuning aligned language models compromises safety, even when users do not intend to!
\newblock In \emph{International Conference on Learning Representations}, volume 2024, pages 30988--31043, 2024.

\bibitem[Rofin et~al.(2026)Rofin, Varre, and Flammarion]{rofin2026learning}
Mark Rofin, Aditya Varre, and Nicolas Flammarion.
\newblock (how) learning rates regulate catastrophic overtraining.
\newblock \emph{arXiv preprint arXiv:2604.13627}, 2026.

\bibitem[Sakaguchi et~al.(2021)Sakaguchi, Bras, Bhagavatula, and Choi]{sakaguchi2021winogrande}
Keisuke Sakaguchi, Ronan~Le Bras, Chandra Bhagavatula, and Yejin Choi.
\newblock Winogrande: An adversarial winograd schema challenge at scale.
\newblock \emph{Communications of the ACM}, 64\penalty0 (9):\penalty0 99--106, 2021.

\bibitem[Sanyal et~al.(2025)Sanyal, Prairie, Das, Kavis, and Sanghavi]{sanyal2025upweighting}
Sunny Sanyal, Hayden Prairie, Rudrajit Das, Ali Kavis, and Sujay Sanghavi.
\newblock Upweighting easy samples in fine-tuning mitigates forgetting.
\newblock \emph{arXiv preprint arXiv:2502.02797}, 2025.

\bibitem[Shenfeld et~al.(2026{\natexlab{a}})Shenfeld, Damani, H{\"u}botter, and Agrawal]{shenfeld2026self}
Idan Shenfeld, Mehul Damani, Jonas H{\"u}botter, and Pulkit Agrawal.
\newblock Self-distillation enables continual learning.
\newblock In \emph{International Conference on Machine Learning}, 2026{\natexlab{a}}.

\bibitem[Shenfeld et~al.(2026{\natexlab{b}})Shenfeld, Pari, and Agrawal]{shenfeld2026rlrazor}
Idan Shenfeld, Jyothish Pari, and Pulkit Agrawal.
\newblock {RL}'s razor: Why online reinforcement learning forgets less.
\newblock In \emph{International Conference on Learning Representations}, 2026{\natexlab{b}}.

\bibitem[Tang et~al.(2023)Tang, Deng, Lin, Han, Liang, Cao, and Sun]{tang2023toolalpaca}
Qiaoyu Tang, Ziliang Deng, Hongyu Lin, Xianpei Han, Qiao Liang, Boxi Cao, and Le~Sun.
\newblock Toolalpaca: Generalized tool learning for language models with 3000 simulated cases.
\newblock \emph{arXiv preprint arXiv:2306.05301}, 2023.

\bibitem[Wu et~al.(2025)Wu, Tam, Lin, Chen, Sun, and Lee]{wu2025mitigating}
Chao-Chung Wu, Zhi~Rui Tam, Chieh-Yen Lin, Yun-Nung Chen, Shao-Hua Sun, and Hung-yi Lee.
\newblock Mitigating forgetting in llm fine-tuning via low-perplexity token learning.
\newblock In \emph{Advances in Neural Information Processing Systems}, volume~38, 2025.

\bibitem[Zellers et~al.(2019)Zellers, Holtzman, Bisk, Farhadi, and Choi]{zellers2019hellaswag}
Rowan Zellers, Ari Holtzman, Yonatan Bisk, Ali Farhadi, and Yejin Choi.
\newblock Hellaswag: Can a machine really finish your sentence?
\newblock In \emph{Proceedings of the 57th annual meeting of the association for computational linguistics}, pages 4791--4800, 2019.

\bibitem[Zheng et~al.(2025)Zheng, Guha, Arifov, Zhang, Skreta, Manning, Henderson, and Ho]{zheng2025reasoning}
Lucia Zheng, Neel Guha, Javokhir Arifov, Sarah Zhang, Michal Skreta, Christopher~D Manning, Peter Henderson, and Daniel~E Ho.
\newblock A reasoning-focused legal retrieval benchmark.
\newblock In \emph{Proceedings of the 2025 Symposium on Computer Science and Law}, pages 169--193, 2025.

\bibitem[Zhou et~al.(2023)Zhou, Lu, Mishra, Brahma, Basu, Luan, Zhou, and Hou]{zhou2023instruction}
Jeffrey Zhou, Tianjian Lu, Swaroop Mishra, Siddhartha Brahma, Sujoy Basu, Yi~Luan, Denny Zhou, and Le~Hou.
\newblock Instruction-following evaluation for large language models.
\newblock \emph{arXiv preprint arXiv:2311.07911}, 2023.

\end{thebibliography}
